\documentclass{CUP-JNL-NLP-arxiv}

\usepackage{amsmath}
\usepackage{graphicx}
\usepackage{natbib}
\ifpdf%
\usepackage{epstopdf}%
\else%
\fi
\usepackage{multirow}

\usepackage{orcidlink}

\usepackage{newfloat}

\usepackage{hyperref}

\usepackage{todonotes}

\usepackage{amssymb}
\usepackage[makeroom]{cancel}

\usepackage{amsthm}

\usepackage{array}
\usepackage{multirow}

\usepackage{caption}
\usepackage{subcaption}

\usepackage[most]{tcolorbox}

\usepackage{placeins}

\usepackage{graphicx} 

\usepackage{pgfplots}
\pgfplotsset{compat=1.18}
\usepackage{tikz}

\usepackage{listings}

\usepackage[
leftmargin=5mm,
rightmargin=4mm,
vskip=1pt
]{quoting}

\usepackage{pgf}

\usepackage[table,xcdraw]{xcolor}

\usepackage[normalem]{ulem}
\usepackage{contour} 

\definecolor{LLMbg}{RGB}{248,248,248}
\definecolor{LLMframe}{RGB}{210,210,210}
\definecolor{LLMcomment}{RGB}{90,90,90}

\newcommand{\cc}[1]{\cellcolor[HTML]{#1}}

\lstdefinelanguage{LLMPrompt}{
  sensitive=true,
  morecomment=[l]{\#},
}

\lstdefinestyle{llmprompt}{
  language=LLMPrompt,
  basicstyle=\ttfamily\small,
  columns=fullflexible,
  keepspaces=true,
  upquote=true,
  backgroundcolor=\color{LLMbg},
  frame=single,
  rulecolor=\color{LLMframe},
  framesep=6pt,
  frameround=tttt,
  xleftmargin=1.5em,
  framexleftmargin=1.2em,
  breaklines=true,
  breakatwhitespace=true,
  prebreak=\mbox{\textcolor{LLMcomment}{\tiny$\hookleftarrow$}},
  postbreak=\mbox{\textcolor{LLMcomment}{\tiny$\rightarrow$}},
  showstringspaces=false,
  numbers=left,
  numberstyle=\tiny\color{LLMcomment},
  numbersep=8pt,
  escapeinside={(*@}{@*)},
  breakindent=1em,
}

\newtcolorbox{quotebox}{
  enhanced,
  breakable,
  colback=white,
  sharp corners,
  frame hidden,
  borderline west={1mm}{0pt}{gray},
  boxsep=0mm,
  left=2mm,
  left skip=.5mm,
  top=.5mm,
  right=1mm,
  bottom=.5mm,
  before=,
  before skip=2mm,
  middle=0mm,
  after=,
  before upper=\color{darkgray},
}

\contourlength{0.8pt}

\newcommand{\myuline}[1]{%
  \uline{\phantom{#1}}%
  \llap{\contour{white}{#1}}%
}

\useunder{\uline}{\ul}{}

\theoremstyle{definition}

\usetikzlibrary{calc}

\pgfplotsset{
    discard if not/.style 2 args={
        x filter/.append code={
            \edef\tempa{\thisrowno{#1}}
            \edef\tempb{#2}
            \ifx\tempa\tempb
            \else
                
            \fi
        }
    }
}

\newcolumntype{A}{ >{$} r <{$} @{} >{${}} l <{$} } 

\newcommand*{\MinNumber}{0}
\newcommand*{\MidNumber}{0.5}
\newcommand*{\MaxNumber}{1}

\definecolor{lightgray2}{RGB}{220,220,220}

\newcommand*{\MinColor}{BurntOrange}
\newcommand*{\MidColor}{lightgray2}
\newcommand*{\MaxColor}{Cyan}

\newcommand{\ApplyGradient}[1]{%
    \getargs[q]{#1}%
    \ifdim \narg pt>0 pt
        \edef\argnumber{\argi}
        \IfDecimal{\argnumber}{
            \ifdim \argnumber pt>\MidNumber pt%
                \pgfmathsetmacro{\PercentColor}{max(min(100.0*(\argnumber-\MidNumber)/(\MaxNumber-\MidNumber),100.0),0.00)}%
                \ifdim \narg pt<2 pt
                    \edef\x{\noexpand\cellcolor{\MaxColor!\PercentColor!\MidColor}}%
                \else
                    \edef\x{\noexpand\cellcolor{\argii}}%
                \fi%
                \x\argnumber%
            \else%
                \pgfmathsetmacro{\PercentColor}{max(min(100.0*(\MidNumber-\argnumber)/(\MidNumber-\MinNumber),100.0),0.00)}%
                \ifdim \narg pt<2 pt
                    \edef\x{\noexpand\cellcolor{\MinColor!\PercentColor!\MidColor}}%
                \else
                    \edef\x{\noexpand\cellcolor{\argii}}%
                \fi%
                \x\argnumber%
            \fi%
        }{\argnumber}%
    \else%
    \fi%
}

\newcolumntype{R}{>{\collectcell\ApplyGradient}c<{\endcollectcell}}

\DeclareCaptionLabelFormat{cont}{#1~#2\alph{ContinuedFloat}}

\makeatletter
\providecommand*\@chapterlistsgap@on[1]{}%
\providecommand*\@chapterlistsgap@off[1]{}%
\makeatother

\DeclareFloatingEnvironment[
    fileext=lopc,                  
    listname={List of Pseudocode}, 
    name=Pseudocode,               
]{pseudocode}

\DeclareFloatingEnvironment[
    fileext=loe,                  
    listname={List of Examples},  
    name=Example,                 
]{example}

\DeclareFloatingEnvironment[
    fileext=lop,                  
    listname={List of Prompts},   
    name=Prompt,                  
    autorefname=Prompt,
]{prompt}

\makeatletter

\newcommand\punctfootnote{%
  \@ifnextchar[{\punctfootnote@i}{\punctfootnote@i[]}%
}
\def\punctfootnote@i[#1]{%
  \@ifnextchar[{\punctfootnote@ii[#1]}{\punctfootnote@ii[#1][0]}%
}
\def\punctfootnote@ii[#1][#2]#3{%
  \begingroup
    \if@minipage
      \ifcsname H@refstepcounter\endcsname
        \H@refstepcounter\@mpfn%
      \else
        \stepcounter\@mpfn%
      \fi
      \protected@xdef\@thefnmark{\thempfn}%
    \else
      \ifcsname H@refstepcounter\endcsname
        \H@refstepcounter{footnote}%
      \else
        \refstepcounter{footnote}%
      \fi
      \protected@xdef\@thefnmark{\thefootnote}%
    \fi
    \def\punctfn@pos{#2}%
    \dimen@=\punctfn@pos\p@\relax
    \ifdim\dimen@<\z@ \def\punctfn@pos{0}\fi
    \ifdim\dimen@>\p@ \def\punctfn@pos{1}\fi
    \let\punctfn@orig@makefnmark\@makefnmark
    \def\punctfn@punct{#1}%
    \def\@makefnmark{%
      \setbox0=\hbox{\@textsuperscript{\normalfont\@thefnmark}}%
      \setbox2=\hbox{\punctfn@punct}%
      \dimen2=\punctfn@pos\wd2\relax
      \dimen4=\wd0 \advance\dimen4 by \dimen2
      \hbox to \dimen4{%
        \rlap{\kern\dimen2 \copy0}%
        \box2\hss
      }%
    }%
    \if@minipage
      \@mpfootnotemark
      \let\@makefnmark\punctfn@orig@makefnmark
      \@mpfootnotetext{#3}%
    \else
      \@footnotemark
      \let\@makefnmark\punctfn@orig@makefnmark
      \@footnotetext{#3}%
    \fi
  \endgroup
}

\renewcommand\footnote{%
  \@ifnextchar[{\footnote@i}{\punctfootnote[]}%
}
\def\footnote@i[#1]{%
  \@ifnextchar[{\footnote@ii[#1]}{\punctfootnote[#1]}%
}
\def\footnote@ii[#1][#2]#3{
  \punctfootnote[#1][#2]{#3}%
}

\makeatother

\begin{document}
\label{firstpage}

\righttitle{Schouten \textit{et al.}}

\papertitle{Article}

\jnlPage{\pageref{firstpage}}{\pageref{lastpage}}
\jnlDoiYr{2026}
\doival{10.1017/xxxxx}

\title{\textbf{ToxiREX}: A Dataset on \textbf{Toxi}c \textbf{RE}asoning in Conte\textbf{X}t}

\begin{authgrp}
\author{
Stefan F. Schouten\orcidlink{0000-0001-9839-9985}, 
Ilia Markov\orcidlink{0000-0001-9533-748X}, 
Piek Vossen\orcidlink{0000-0002-6238-5941}
}
\affiliation{
    Vrije Universiteit Amsterdam, the Netherlands\\
    \textbf{Corresponding author: } Stefan F. Schouten;
    Email: \email{s.f.schouten@vu.nl}
}
\end{authgrp}


\begin{abstract}
We introduce a new, contextual, multilingual dataset called ToxiREX: \textbf{Toxi}c \textbf{RE}asoning in Conte\textbf{X}t.
The dataset consists of threads of Reddit comments and structured characterizations of what the comments imply, following a systematic toxic reasoning schema developed in a previous paper \citep{schouten_position_2026}.
Using the schema allows us to capture and explain implicit and context-dependent toxicity, while supporting mappings to existing toxicity taxonomies.
The dataset includes comments in six languages (English, Arabic, Turkish, Spanish, German, and Dutch), collected from posts connected to specific major events (e.g. the 2023 Turkey earthquakes; the Russian invasion of Ukraine).
We describe the context-preserving preprocessing of the threads.
We create a training set of 125 thousand comments which is annotated by a commercially available LLM, and a test set of just under three thousand comments that is annotated by native speakers.
We show that apparent disagreements in the test set annotations often reflect defensible alternative interpretations rather than noise. 
Finally, we provide baseline results by prompting and fine-tuning language models. 
To produce these results, we develop evaluation strategies for our hierarchical, schema-based predictions. 
While models perform better than random, there remains a lot of room for improvement, showing the task to be challenging.
ToxiREX is the first dataset to simultaneously incorporate multiple languages, conversational context, and implicit toxicity, while using the toxic reasoning schema for rich, structured annotations. 
Dataset available at: \url{https://github.com/cltl/toxirex}

{\vspace{3mm}\noindent\color{red}Warning: This paper contains examples of toxic language.}
\end{abstract}
\maketitle
\section{Introduction}
Toxic language, defined here broadly as any language that conveys hateful, derogatory, or offensive ideas \citep{garg_handling_2023}, is widely regarded as undesirable and must be identified and addressed to foster healthier online environments \citep{fortuna_survey_2018}. 
While explicit expressions of toxicity are often easier to recognize, toxicity conveyed implicitly \citep{waseem_understanding_2017} presents a greater challenge \citep{caselli_i_2020}. 
Determining whether speech is implicitly toxic can be subjective, as individuals may interpret the same text in different ways. 
Moreover, understanding the author's intended implications may depend on access to the relevant context, including cultural, situational, and discourse context.

The widespread adoption of social media has significantly amplified the need to automate the detection of toxic language. 
Such language can inflict harm on individuals and communities, reinforce harmful stereotypes, and can contribute to violence \citep{madriaza_exposure_2025}. 
As a result, identifying and mitigating toxic content is essential for maintaining inclusive and healthy online environments.

In this work, we introduce ToxiREX, a dataset of social media comments taken from Reddit. 
These comments have been annotated for toxic implications with the original conversational context present.
For the annotation, we adopt the \textit{toxic reasoning schema} \citep{schouten_position_2026}.
This schema systematizes categories of (implicit) toxic language by breaking down implications into independent traits. 
In doing so, we separate the propositional content of an implication from the attitudes held toward it by various stakeholders.
For three traits in the schema, we also produce span-level annotations to identify the relevant parts of a comment.
We use GPT4o to produce silver labels for each comment in the training data, and organized an annotation campaign to create the test set. 
To establish its quality, we also performed an analysis of apparent disagreements in the annotations of the test set.

The schema supports mapping to and from existing taxonomies of toxic language.
This allows us to envision at least two strategies: 
(1) a bottom-up approach, where data is annotated according to the schema directly and then mapped onto categories of toxic language,
or
(2) a top-down approach, where data is first annotated according to conventional taxonomies, and then mapped onto the schema.
These methods are complementary and both come with their own up- and downsides. 
The top-down approach is straightforward to execute, and annotation campaigns remain relatively simple.
However, the bottom-up approach, while labor-intensive, allows for an open-ended exploration where annotations cover all kinds of potentially toxic content, even that which may not be covered by existing taxonomies.

We adopt the bottom-up approach to develop the ToxiREX dataset, with the goal of investigating:
(1) the feasibility of directly annotating comments according to the toxic reasoning schema;
(2) if the toxic reasoning schema is expressive enough to cover most potentially toxic implications; and finally,
(3) whether existing approaches, like prompting and fine-tuning, can be used to detect and categorize implications according to the toxic reasoning schema.


We find that annotating according to the toxic reasoning is feasible, albeit with some unique challenges.
Because of the hierarchical nature of the schema, the annotations are more sparse, making it harder to calculate to what degree annotators agreed.
Similarly, this also complicates evaluating model predictions.
The findings of the annotation campaign suggest the toxic reasoning schema is able to characterize a wide variety of potentially toxic implications expressed in many different ways in different languages.

Finally, our experiments show that detecting and characterizing the implications of potentially toxic comments is a very challenging task.
While LLMs perform reasonably well when evaluated in a zero-shot setting, there remains substantial room for improvement.
Fine-tuning a smaller language model on LLM predictions retains much of the LLM's performance at a much smaller computational budget.

\section{Related Work}
Toxic language detection has been studied extensively in the last decade \citep{jahan_systematic_2023,fortuna_survey_2018}.
Here we compare and contrast our contribution to previous work along various dimensions.

\paragraph{Fine-grained classification}
Toxic language detection is often approached as a binary classification problem, but others have broken down the problem by introducing fine-grained categorizations. 
For example, \citet{vidgen_introducing_2021}, \citet{elsherief_latent_2021}, \citet{kirk_semeval-2023_2023}, and \citet{pachinger_disaggregated_2025} all  introduce (multi-level) taxonomies for variants of toxic language detection.
We also break down the problem, but rather than a hierarchy, we use the toxic reasoning schema, which has many independent traits. 
Simultaneously, specific combinations of these traits can still be mapped onto existing categories of toxic language, when desired.

\paragraph{Span-level and Structured annotation}
Various datasets have introduced span-level annotations. 
Some providing general `toxic spans', which indicate a particular span has a higher degree of responsibility towards the toxicity than the rest of the comment \citep{pavlopoulos_semeval-2021_2021,mathew_hatexplain_2021}.
Others include spans that indicate more specific aspects, such as the target of the toxicity \citep{barbarestani_annotating_2022,jafari_target_2024}.
Our work similarly includes spans that mark the target, and optionally spans marking a group or individual to which the target is compared.

Previous work has pointed out that human moderators use moderation guidelines to decide if posts are problematic \citep{calabrese_explainable_2022}.
They argue that many cases depend on the exact rules/norms that should be applied.
Motivated by this they use intent classification and slot filling (ICSF), which requires parts of the text (spans) beyond merely the target to be assigned to sets of possible slots. 
A similar work is that of \citet{pachinger_disaggregated_2025}, which similarly annotates multiple kinds of spans.
Our work also addresses the dependency of toxic language detection on guidelines, and similarly uses structured multiple-span annotation.
But, instead of designing slots for a specific guideline, we use the toxic reasoning schema, which is intended to generalize across guidelines.

\paragraph{Multilinguality}
Many existing toxic language datasets are monolingual, with most of the resources being in English.
However, previous work has pointed out the importance of having resources that cover multiple languages \citep{basile_semeval-2019_2019}.
Existing multilingual resources include \citet{ousidhoum_multilingual_2019} and \citet{yadav_lahm_2022}.
Our work incorporates six languages: Arabic, Dutch, English, German, Spanish, and Turkish.
Our dataset further distinguishes itself by connecting comments and posts to major world event across different language-cultural groups. 
Because of this connection, the data uniquely enables the investigation of variation in toxic language reasoning and expression across these communities, as well as the capacity of LLMs to reflect these.

\paragraph{Implicit Toxicity}
In the literature, a distinction is often made between explicit and implicit toxicity.
The most explicit forms of toxicity involve the use of words which are almost always toxic.
Implicit toxicity often comes in two degrees: (1) people plainly asserting something harmful or offensive, but without using toxic words; and (2) people stating one thing while implying another.
Recent work primarily targeting the latter forms of implicit toxicity has done so by including free-text annotations of implied statements \citep{sap_social_2020,elsherief_latent_2021}.
We also include free-text annotations of implied statements, but further characterize the implications using the traits of the toxic reasoning schema.

\paragraph{In-Context Annotation}
Many existing datasets of toxic language include social media posts on their own, with no context \citep[e.g.][]{basile_semeval-2019_2019,pavlopoulos_semeval-2021_2021}.
However, often posts are part of conversations, and cannot be understood independently.
Previous work has firmly established that conversational context is important to understand if and why a post is toxic \citep{pavlopoulos_toxicity_2020,yu_hate_2022,xenos_toxicity_2022,ljubesic_quantifying_2023}.
The comments in our dataset are all annotated within their original conversational context i.e. the Reddit thread they were posted in.

\par\vspace{3.25ex plus 1ex minus .2ex}%
\noindent%
While there are existing datasets that have one or some of the features mentioned above, to the best of our knowledge, ToxiREX is the first to combine all of them.

\section{Toxic Reasoning}\label{sec:toxic_reasoning_summary}
In \citet{schouten_position_2026}, we defined toxic reasoning as ``toxic language detection that further requires making explicit the reasons for why something should be considered toxic'' and proposed that it may be approached with the toxic reasoning schema. 
In this schema, what is conveyed by a toxic message is characterized in two parts: 
(1) a proposition predicating something of a subject, and
(2) the author (implicitly) reporting an attitude (of belief, desire, or intent) towards this proposition. 
The subject will often be the target of the toxicity (when expressing negativity) , but can also be the author's in-group (when expressing exclusionary or discriminatory positivity). 

Besides the author's attitudes, we also include the attitudes of the general public and experts\footnote{A person with the expertise to assess the truth of the relevant proposition.} toward the proposition, which allows us to further differentiate between specific kinds of toxicity. 
The schema is used along with mappings from the schema to existing taxonomies, like that of the Implicit Hate Corpus \citep[IHC,][]{elsherief_latent_2021}.
Such mappings allow for schema-based annotations and predictions to be converted back to (fine-grained) classifications.

\subsection{Toxic Reasoning Schema}\label{ssec:schema_summary}
Here we will briefly summarize the toxic reasoning schema. 
The schema's first trait categorizes what is at the core of a message's implication: 
the message {content} characterizes the proposition that is central to what the text conveys. 
The schema distinguishes between three high-level \textbf{categories} of content.%
\begin{itemize}%
    \item{}$\textit{Situation}(\texttt{subject}, [\texttt{other}])$: 
    a situation (e.g. environment, condition, circumstance, etc.) applies to the \texttt{subject}.  
    \item{}$\textit{Quality}(\texttt{subject}, [\texttt{other}])$:  
    the \texttt{subject} possesses a given inherent quality, or has a certain nature.
    \item{}$\textit{Behavior}(\texttt{subject}, [\texttt{other}])$: 
    the \texttt{subject} behaves in a particular way.
\end{itemize}
Each can occur with \textit{Negative} or \textit{Positive} \textbf{polarity}, and can optionally apply relative to a an `other' (group or individual).

A number of other traits are included in the schema.
For example, IHC's `Threatening and Intimidation' is mapped onto the schema as follows.
\citet{elsherief_latent_2021} state that this class ``convey[s] a speaker commitment to a target’s pain, injury, damage, loss, or violation of rights.'', explicitly choosing a wider definition that does not focus solely on threatening `life or limb'. 
Thus, we say that the class applies if and only if it expresses an intention to subject (\textit{future}) a person or group of people to a \textit{Negative} \textit{Situation}.
The \textit{future}-oriented nature of threats is covered by \textbf{temporality} trait.

Now consider `Inferiority language', which can, for example, include instances where someone states or implies that a \textit{group} has a \textit{Positive} \textit{Quality} or \textit{Behavior} (e.g. ``Japanese people have higher IQ'',  or ``white people invented civilization'').
In this case, in order for it to be considered toxic, we also require that an expert has an attitude of disbelief towards the message's content.
The fact that these instances necessarily involve a group, is covered by the \textbf{specificity} trait.

The schema includes a few other traits, including the \textbf{group-type}, and the \textbf{subject-role} / \textbf{other-role}, which specify the role of subject in the conversation, i.e. is the subject or other: the author, another conversation participant, someone else, or a group that includes them.
For a complete description of the schema, see \citet{schouten_position_2026}. 

Finally, the schema supports identifying spans that clarify which part of a message is responsible for each of these traits.
In ToxiREX, we include spans highlighting the \texttt{subject}, the \texttt{other}, and the content \textbf{category}.

\FloatBarrier

\begin{figure}
    \centering%
    \includegraphics[width=0.45\linewidth]{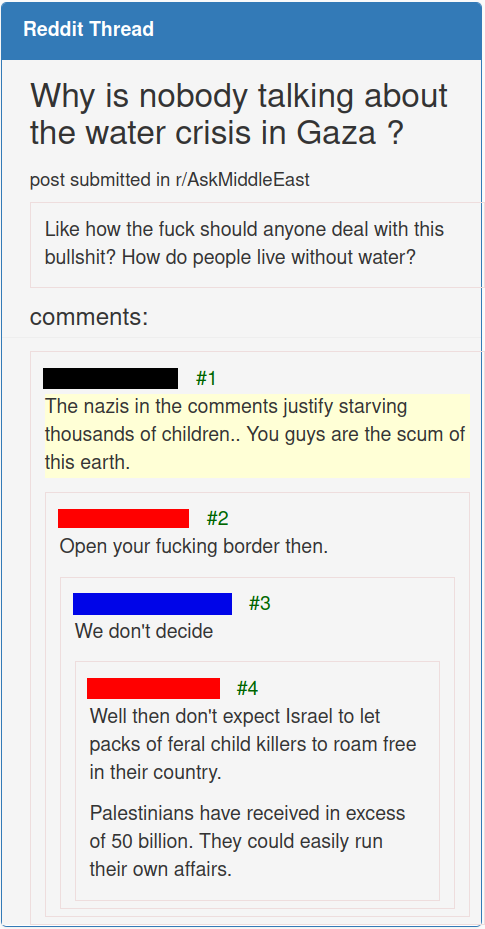}%
    \caption{Thread of comments as presented to annotators. User names have been redacted by replacing each name with a colored box.}%
    \label{ex:thread}%
\end{figure}

\section{Data Selection \& Pre-processing}
\begin{figure}
    \centering
    \includegraphics[width=0.8\linewidth]{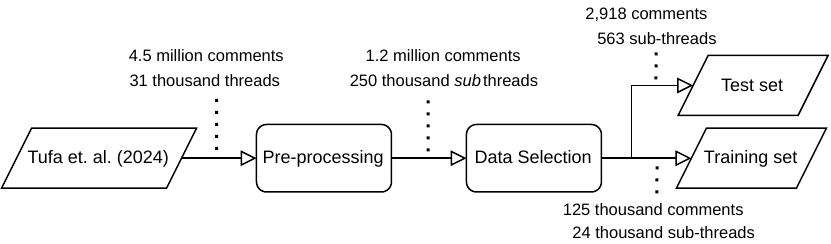}
    \caption{Flow chart showing the data processing steps.}
    \label{fig:flow_chart}
\end{figure}

In this section, we give details on the following:
(1) the source data,
(2) context-preserving pre-processing of the data; and,
(3) selection of a subset which contains sufficient and diverse toxicity.
See \autoref{fig:flow_chart} for an overview.

\subsection{Source Data}
Rather than collect data from scratch, we use the Reddit data scraped by \citet{tufa_grounding_2024}, but with our own pre-processing.
This data was selected for its relation to one of fifteen major real-world events spanning four years from 2020 to 2023. 
Events cover a wide range of topics, for example: the January 6th U.S. capitol attack; the 2023 Turkey earthquakes; the Russian invasion of Ukraine, etc%
. (See \autoref{apx:toxirex_events} for the full list.)
The raw data consists of just under 4.5 million \textit{comments} from 31 thousand \textit{posts} in six languages: English (EN), Dutch (NL), Arabic (AR), Turkish (TR), Spanish (ES), and German (DE).

On Reddit, comments can be left directly in response to posts, or comments can reply to another comment, forming a tree structure.
An example of some comments can be seen in \autoref{ex:thread}.

\subsection{Pre-processing}\label{sssec:pre-processing}
Our data pre-pro\-cess\-ing is designed to allow studying toxicity in context, while maintaining annotation feasibility. 
While it would be optimal to annotate all comments in their full context, in practice asking annotators to read up to dozens of comments is not practical.
So, we need to select data where the context is of a manageable size. 
A naive approach could be to select only comments up to a certain depth, but that would involve throwing away lots of potentially interesting data. 
Thus, we make a compromise by splitting up the comments below a post into what we call \textit{sub-threads}:  sequences of comments where each comment is a reply to the previous.

The sub-thread creation process for a given post proceeds as follows:
\begin{enumerate}
    \item Each comment is associated with its `ancestry', i.e. its parent\punctfootnote[,][0.5]{A comment's parent is the comment it replies to.} its parent's parent, and so on until the top-level. 
    \item The candidate sub-threads consist of every leaf comment's ancestry\punctfootnote[.][0.5]{A leaf comment is a comment without replies.}
    \item Leaf comments whose ancestry is shorter than the minimum sub-thread length of three comments are dropped from consideration.
    \item From among the remaining candidate sub-threads, we now look for a candidate $C$ to be confirmed as a sub-thread. 
    This selection is done by heuristic. 
    We look for the candidate that minimizes the following formula: $|L_C - L_\text{ideal}| + B_C$, where $L_C$ is the candidate's length, $L_\text{ideal} = 5$ is the ideal length we choose, and $B_C$ is the number of branching points in $C$, i.e. the number of comments that have more than one reply.
    The goal is to prioritize candidates close to the ideal length, that also leave the tree as intact as possible.
    \item The confirmed sub-thread's comments are now removed from further consideration and removed from the ancestries of the remaining comments. 
    We now return to step 2, and the process is repeated until no candidates remain.
\end{enumerate}

\begin{figure}
    \centering
    \includegraphics[width=0.5\linewidth]{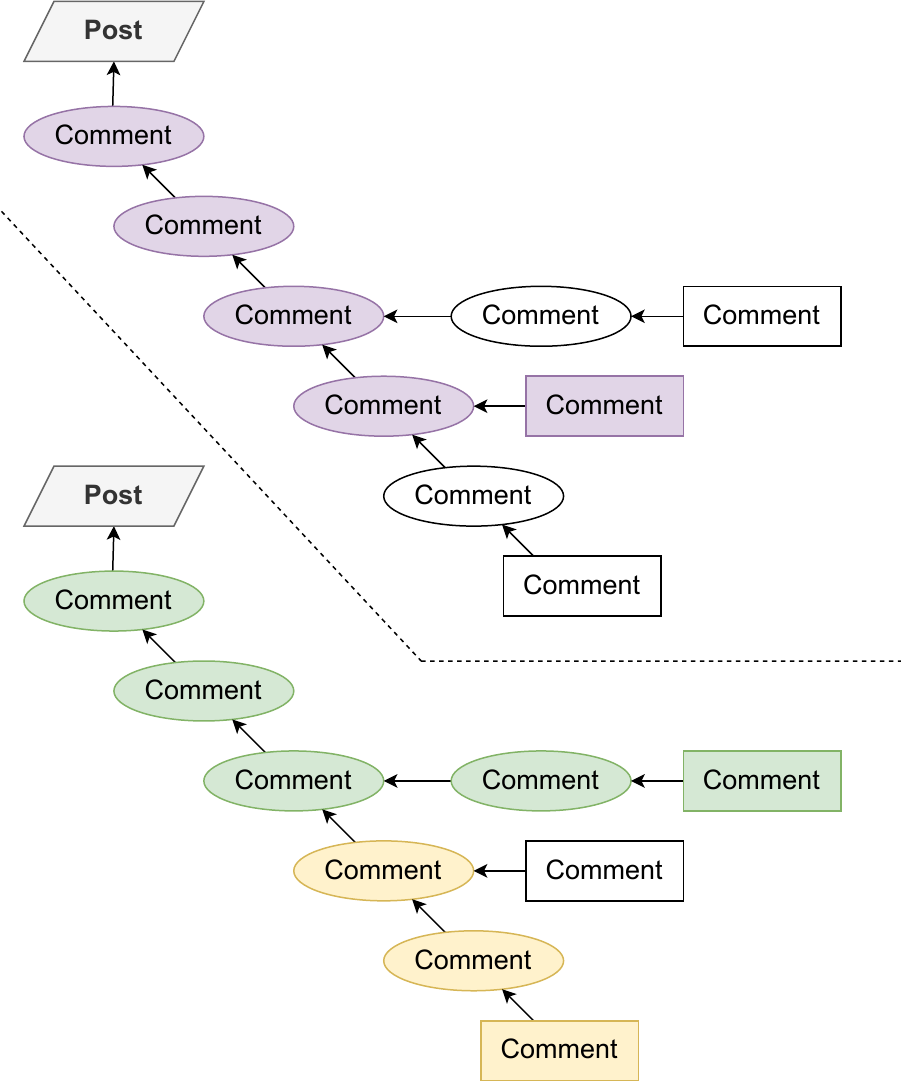}
    \caption{Example of sub-thread creation from a post's comment hierarchy. On the top, it shows an example of creating a sub-thread (purple) with two branching points, it leaves four unusable comments. On the bottom, it shows how selecting the candidate with fewer branching points (green), leaves another viable sub-thread (yellow).}
    \label{fig:sub-thread-example}
\end{figure}

\autoref{fig:sub-thread-example} depicts a small example comment hierarchy that demonstrates the motivation behind the heuristic.

\subsection{Data Selection}\label{ssec:data_selection}
We use an LLM to select data in two steps.

\paragraph{Step 1}
We suspect that the prevalence of toxic language in the full set of sub-threads is fairly low.
Therefore, we use an LLM to clasify sub-threads by their toxicity, and GPT4o finds around 10\% of the sub-threads to be toxic. 
So, rather than annotate everything, we use this subset for the dataset.
Knowing that LLMs tend to be biased toward classifying text as toxic \citep{kumar_watch_2024}, we assume that doing so will keep most toxicity, including implicit toxicity, while removing sub-threads that are clearly non-toxic. 
The prompt we used can be seen in \autoref{prompt:hate_speech}.

\begin{prompt}
\begin{lstlisting}[style=llmprompt]
Read the title: {title}
Read the post text: {post_text}
Read the full conversation: {context}
Note that the name of the author of each comment in the conversation is written at the start of the comment in brackets []
Read the comment: {comment}
Note that this comment is one of the comments in the conversation.
Given all provided information, is the comment an instance of hate speech? Say True if it is an instance of hate speech. Say False if it is not an instance of hate speech.
\end{lstlisting}%
\vspace{-2mm}%
\caption{Used for selecting comments likely to be toxic.}%
\label{prompt:hate_speech}%
\end{prompt}

\paragraph{Step 2}
The goal of the human test set annotation is validate the expressivity of the toxic reasoning schema, and to produce a high-quality test set.
For this purpose, we first perform another selection, this time to create a stratified sample. 
Specifically, we ask an LLM to categorize comments into a few high-level categories: Harmed, Inferior, Negative, NonHuman, Stereotype, Superior\punctfootnote[.][0.5]{These categories are based on an early version of the toxic reasoning schema (TRS0). See Appendix \ref{apx:trs0} and \ref{apx:pydantic_trs0} for details.}

See \autoref{tab:category_stats} for the statistics of how often each category was assigned to the data by the LLM, and exactly how many we sampled from each.
Note that we did not sample the exact same number of comments from each combination of category and language. 
This is because categories apply on the comment level, but we sample entire sub-threads, which also contain comments of the other categories. 
This makes an exact uniform sampling over the categories difficult to realize.

The data which was not selected for human annotation was annotated by OpenAI's GPT4o and used as training data in our experiments.

\begin{table}
\small
\centering
\caption{Number of sub-threads per language and TRS0 category in pre-annotation selection. Each cell shows nr. of sub-threads in test set out of total nr. of sub-threads.}
\label{tab:category_stats}
\let\mc\multicolumn
\begin{tabular}{@{}lr@{ / }lr@{ / }lr@{ / }lr@{ / }lr@{ / }lr@{ / }l@{}}
\toprule
                  & \mc{2}{c}{English} & \mc{2}{c}{Spanish} & \mc{2}{c}{German} & \mc{2}{c}{Turkish} & \mc{2}{c}{Dutch} & \mc{2}{c}{Arabic} \\
\midrule
Harmed            & 20   & 5337     & 23   & 804      & 34    & 449     & 23    & 298     & 25    & 210     & 35    & 130     \\
NotHuman          & 24   & 2185     & 31   & 444      & 22    & 207     & 21    & 147     & 21    & 101     & 30    & 81      \\
Superior          & 24   & 1129     & 33   & 245      & 36    & 111     & 21    & 68      & 7     & 50      & 26    & 46      \\
Inferior          & 25   & 3230     & 12   & 875      & 19    & 293     & 19    & 237     & 14    & 131     & 34    & 117     \\
KnownStereotype   & 24   & 3018     & 27   & 641      & 28    & 340     & 35    & 91      & 29    & 138     & 18    & 32      \\
SomethingNegative & 14   & 10863    & 21   & 2120     & 27    & 934     & 24    & 720     & 23    & 480     & 44    & 290     \\
SomethingPositive & 7    & 1514     & 6    & 326      & 7     & 178     & 19    & 146     & 6     & 81      & 20    & 93      \\
\midrule
Nr. Sub-threads   & 102  & {17935}  & 95   & {3431}   & 98    & {1518}  & 97    & {1026}  & 79    & {691}   & 92    & {388}   \\
Nr. Comments      & 499  & {92139}  & 504  & {17600}  & 497   & {7635}  & 503   & {5352}  & 419   & {3668}  & 496   & {2106}  \\
\bottomrule
\end{tabular}
\end{table}

\begin{figure}
    \centering
    \includegraphics[width=.6\linewidth]{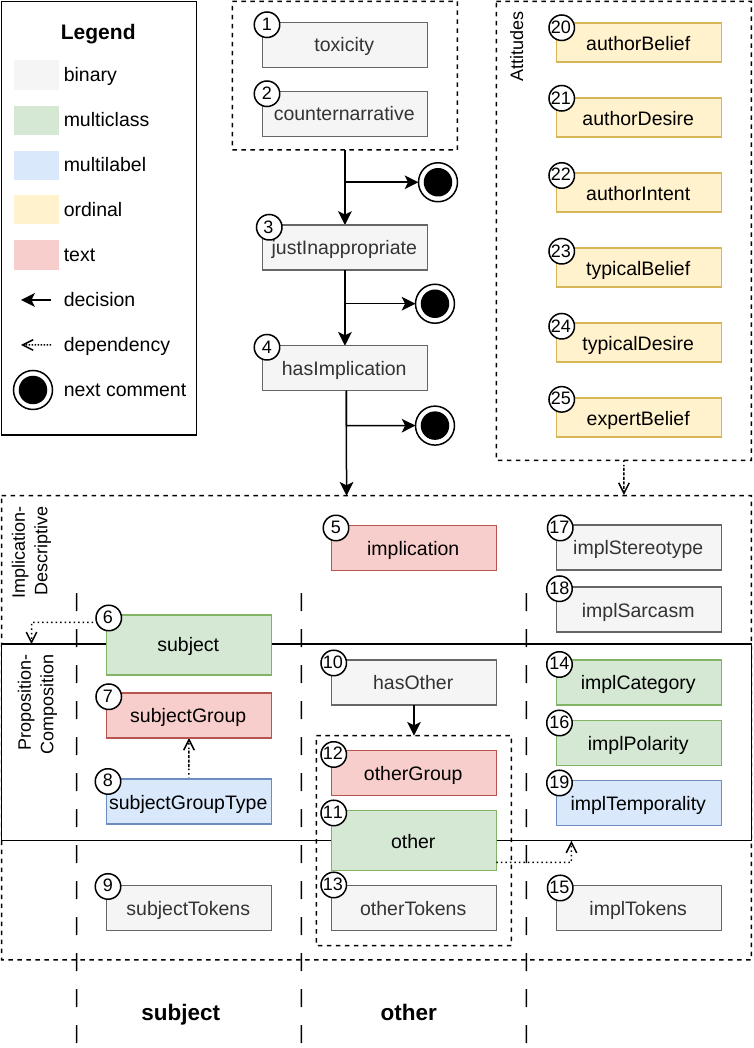}
    \caption{Schematic overview of variables in the annotation. The numbers in the top-left corners of each box indicate the order in which the variables were asked about. Arrows show decisions made during the annotation (solid) and dependencies between variables (dotted).}
    \label{fig:annotation_variable_schema}
\end{figure}

\section{Test Data Annotation}
Here, we will present the results of annotating data according to the toxic reasoning schema. 

Annotators were shown a reproduction of the data as they could have seen it on Reddit (see \autoref{ex:thread}).
This interface was created using Lingoturk\punctfootnote[.][.5]{\url{https://github.com/FlorianPusse/Lingoturk/}}

First, they would be shown the post text, if available\punctfootnote[.][0.5]{Posts on Reddit can also be what are known as `Link Posts', which provide a topic of conversation purely by linking to another webpage, rather than through a written post.}
Next, they would be shown a selection of comments left by Reddit users in response to that post.
Specifically, they would be shown one sub-thread, constructed as described in \ref{sssec:pre-processing}.
One of the comments in that sub-thread would be highlighted, indicating the comment that is to be annotated.
Next, the annotator was shown a number of questions about the comment that was to be annotated.

\subsection{Annotation Details}

\paragraph{Variables}
Each question asked of the annotator corresponds to a variable in \autoref{fig:annotation_variable_schema}. 
The purpose of the first few variables is to prevent annotators from having to do a full annotation for comments that are either clearly and uncontroversially non-toxic, or toxic but merely `inappropriate'.
The answer given to these first questions determines whether or not to fully annotate the comment. This is indicated in \autoref{fig:annotation_variable_schema} with a solid black arrow.

The binary \texttt{toxicity} variable should equal `Yes' if the annotator believes the ``comment might be (perceived as) potentially toxic, polarizing, or otherwise negative toward a group or individual''. 
The \texttt{counternarrative} variable should equal `Yes' if the comment is ``an argument-based reply aimed at the previous comment, providing an alternative perspective''\punctfootnote[.][.5]{Definition in line with previous work \citep{carthy_counter-narratives_2020}.}
If the answer is `No' for both of the above, the annotation proceeds with the next comment.

The \texttt{justInappropriate} variable must be `Yes' if the comment is ``only toxic because it uses inappropriate words, e.g. use of slurs, name-calling, etc.'', with no toxic implication whatsoever.
If the answer is `Yes', the annotation proceeds with the next comment.

Finally, the \texttt{hasImplication} field asks annotators to confirm whether they have identified a toxic implication. 
If, despite making it past the previous variables, the annotator cannot identify what toxicity the comment is communicating (explicitly or implicitly), the annotator can answer `No' and the annotation proceeds with the next comment.

We added \texttt{implication}, which is a textual description of the main implication.
Annotators are asked to identify what they believe to be the main toxic implication of the comment.
The rest of the questions are asked with respect to whatever implication is chosen by the annotator.
The primary aim of including the \texttt{implication} is to encourage the annotator to actually specify an implication before proceeding. 
For the rest of the annotation, the \texttt{implication} remained at the top of the screen for reference. 

The rest of the variables are based on traits of the toxic reasoning schema. 
Typically, the correspondence is one-to-one. 
The only exceptions are the \texttt{subject} and \texttt{other} variables, which include aspects of both the \textit{role} and \textit{specificity} traits.
See \autoref{sec:annotation_questions} for an overview of all questions.

In addition to those variables, we also added the \texttt{subjectTokens}, \texttt{otherTokens} and \texttt{implTokens}, which asked annotators to indicate which tokens in the comment (or its context) corresponded to the \texttt{subject}, \texttt{other} and \texttt{implTopic}, respectively.

\paragraph{Summary sentence}
Another part of the annotation tool was the \textit{summary sentence}.
This is a sentence generated with a simple template from the labels chosen by the annotator. 
Looking at \autoref{fig:annotation_variable_schema}, this sentence was shown after the variables that are part of the box labeled `Implication', but before the `Attitudes'.
The summary sentence uses the variables under `Proposition-Composition' to fill out a sentence template that creates an English sentence. 

For example, for the following comment from \citet{yu_hate_2022}:
\begin{quotebox}%
\myuline{That chick$_\textsc{subj}$} needs a high-five in the face with a chair. Damn \myuline{her$_\textsc{subj}$} for making us look bad!
\end{quotebox}
For an annotation of: \texttt{Harm}, temporality of \textit{future}, specificity \textit{personal}, and subject role \textit{other}; the summary sentence would be: 
``An individual [`That', `chick', `her'] will have harm come to them.''

These sentences were presented to the annotators, who were asked to annotate the attitudes with respect to this sentence.
For example, for the \textsl{author\_desire} trait, we asked:  ``Based on the comment, to what degree does the author seem to wish (or prefer) that: SUMMARY\_SENTENCE?''.
See \autoref{apx:summary_sentence_pseudocode} for pseudocode for how such sentences can be generated.

\paragraph{Annotators}
For each language, three annotators who spoke the language fluently or natively completed the annotation for each comment.
The English and Dutch comments were annotated by experts, while for the other four languages, students were recruited to perform the annotation.
The expert annotators did various small trial annotations of five threads each, which were also used to refine the software with which the annotation was performed.

Two batches of five English threads were selected to train the student annotators in two stages.
Each stage was followed by a meeting where feedback was presented and opportunity provided for the students to ask questions.
At every point in the process, the students were given the opportunity to stop annotating if the exposure to toxic content was too much.

To get an indication of the biases annotators may have, we asked them to share some information about themselves. 
Following the recommendations of \citet{vidgen_directions_2020}, we report demographics, as well as information on (level of) expertise and personal experiences with toxic language. 
This information can be seen in \autoref{apx:annotator_information}, \autoref{tab:annotator_experience} and \ref{tab:annotator_info_A}.


\begin{table}
\setlength{\tabcolsep}{3pt}%
\small%
\caption{A full (sub-)thread from ToxiREX.}%
\label{tab:example_thread}%
\vspace{-2mm}%
\begin{subtable}{\textwidth}
\centering%
\subcaption{Comments and consensus annotation (based on value for first 4 variables)}%
\begin{tabular}{@{}lm{9cm}m{16mm}@{}}
\toprule
    \textbf{\#} & \textbf{comment} & \textbf{annotation} \\
\midrule
    A1. &
    \begin{tabular}[c]{@{}m{9cm}@{}}%
        Why do the Ukrainians think it’s okay to participate in a protest while they’re in a country that isnt theirs?\\ 
        If they have Polish citizenship its one thing but if they’re only there while the war is going on why should they be protesting anything?
    \end{tabular} 
    & {Impl.}  \\
\midrule
    B1. &
    Maybe they think human rights are for all humans   
    & Countersp.\\
\midrule
    A2. &
    Why does a foreigner have a right to tell a country that is not his own how it should be run, as long as it isn’t a threat to his own?
    & {Impl.} \\
\midrule
    B2. &
    So if you let me into your house and I saw you raping your sister I should just keep my mouth shut because it's your house and your sister?
    & Countersp.\\
\midrule
    A3. &
    Ah yes, stopping what is universally recognized as a crime is imperialism
    & Non-toxic \\
\midrule
    B3. &
    Don't intentionally miss the point. Should I keep my mouth shut because it's your house or should I stand against something that is wrong?
    & Countersp.\\
\midrule
    A4. &
    \begin{tabular}[c]{@{}m{9cm}@{}}%
        I answered you. \\ 
        One situation is an obvious universally accepted crime against humanity.\\ 
        The other is cultural imperialism
    \end{tabular}
    & {Impl.} \\
\midrule
    B4. &
    Hate to break it to ya bud but rape isn't universally accepted as a crime against humanity. In fact, there are instances of prepubescent girls being raped by much, much older men only to then have those girls murdered by way of honor killing. Still want to sit on your hands and cry *cultural imperialism*?
    & Countersp. \\
\midrule
    A5. &
    \begin{tabular}[c]{@{}m{9cm}@{}}%
        You seem very hellbent on justifying rape and incest.\\ 
        But I’ll give it to you, sometimes you are justified in stopping practices that are accepted in another culture.\\ 
        But this isn’t one of those times. Gay marriage is hotly contested in many countries, and it isn’t the place of a foreigner to try and influence the decisions another country makes on some issues
    \end{tabular}
    & {Impl.} \\
\midrule
    B5. &
    Sounds like you actually don't give a shit about human rights or maybe you're just homophobic and are trying to weasel out of coming across as such. This will be my last response because you're just not someone worth interacting with.
    & {Impl.} \\
\bottomrule
\end{tabular}
\end{subtable}
\end{table}

\begin{table}%
\small%
\setlength{\tabcolsep}{3pt}%
\ContinuedFloat%
\vspace{-2mm}%
\begin{subtable}{\textwidth}
\centering%
\subcaption{Comment implications. Square brackets contain \texttt{implTokens} and, if an indivual is targeted, \texttt{subjTokens}.}%
\label{tab:example_thread_A}%
\begin{tabular}{@{}lm{84mm}m{23mm}@{}}
\toprule
    \textbf{\#} & \textbf{implication} & \textbf{author attitude} \\
\midrule
     A1. & Ukrainian refugees in Poland have/will have Negative behavior [``participate in a protest while they're in a country that isnt theirs''] compared to Polish citizens. 
        & \mbox{~~~~\textsl{belief=yes}} \mbox{~~~~\textsl{desire=no}} \\
\addlinespace[2mm]
     A2. & \multicolumn{1}{c}{\textquotedbl} & \multicolumn{1}{c}{\textquotedbl} \\
\addlinespace[2mm]
     A4. & Ukrainians have Negative behavior [``cultural imperialism''].
        & \mbox{~~~~\textsl{belief=yes}} \mbox{~~~~\textsl{desire=no}} \\
\addlinespace[2mm]
     A5. & An individual [``You'', ``you''] has Negative behavior [``justifying rape and incest''].  
        & ~~~~\textsl{belief=yes} \\
\addlinespace[2mm]
     B5. & An individual [``you'', ``you're''] has Negative behavior [``don't give a shit about human rights'', ``homophobic''].  
        & ~~~~\textsl{belief=yes} \\
\bottomrule
\end{tabular}
\end{subtable}
\end{table}

\begin{table}
\small
\centering
\let\mc\multicolumn
\newcommand{\dd}{$^\dagger$}
\caption{Test set statistics, including label statistics for the human annotations. The first three rows give sub-thread lengths. The `Nr. of non-toxic' vs. `possibly toxic' are based on a majority vote. The rest of the rows show the raw number of annotations which has that category or polarity.}
\label{tab:test_set_statistics}
\begin{tabular}{@{}l@{\hspace{1mm}}lrrrrrr@{}}
\toprule
 & & \mc{1}{l}{\textbf{AR}} & \mc{1}{l}{\textbf{DE}} & \mc{1}{l}{\textbf{EN}} & \mc{1}{l}{\textbf{ES}} & \mc{1}{l}{\textbf{NL}} & \mc{1}{l}{\textbf{TR}} \\
\midrule
    & Avg. sub-thread length       & 5.50  & 5.22  & 5.10  & 5.44  & 5.30  & 5.23  \\
    & Min. sub-thread length       & 4     & 4     & 4     & 4     & 4     & 4     \\
    & Max. sub-thread length       & 10    & 10    & 10    & 10    & 9     & 10    \\
\midrule
    & Nr. of non-toxic            & 293   & 246   & 309   & 232   & 265   & 233   \\
    & Nr. of possibly toxic       & 213   & 260   & 211   & 263   & 136   & 217   \\
\midrule
\multirow{7}{*}{\rotatebox{90}{\textbf{Category}}}
    & Situational                 & 14    & 20    & 43    & 45    & 16    & 9     \\
    & ~~Harm                      & 11    & 11    & 30    & 29    & 23    &       \\
    & Qualitative                 & 54    & 64    & 65    & 118   & 24    & 40    \\
    & ~~Dehumanization            & 8     & 4     & 3     & 26    & 3     & 13    \\
    & Behavioral                  & 122   & 96    & 201   & 141   & 149   & 78    \\
    & NonSpecific\dd              & 7     & 2     & 9     & 3     & 3     & 5     \\
    & UnclearOrNone\dd            & 11    & 9     & 6     & 20    & 3     & 3     \\
\midrule
\multirow{4}{*}{\rotatebox{90}{\textbf{Polarity}}}
    & Positive                    & 10    & 8     & 22    & 8     & 12    &       \\
    & Neutral                     & 13    & 6     & 74    & 21    & 10    & 6     \\
    & Negative                    & 157   & 160   & 219   & 275   & 169   & 115   \\
    & None\dd                     &       &       & 2     & 3     & 2     &       \\
\midrule
    & \dd Nr. of problematic      & 18    & 11    & 17    & 26    & 8     & 8     \\
\bottomrule
\\
\toprule
\multirow{11}{*}{\rotatebox{90}{\textbf{Polarity~+~Category}}}
    & Positive-Situational        & 1     & 2     & 5     & 1     & 5     &       \\
    & Neutral-Situational         & 1     & 2     & 9     & 3     &       &       \\
    & Negative-Situational        & 12    & 17    & 29    & 41    & 11    & 9     \\
    & ~~Harm                      & 11    & 11    & 30    & 29    & 23    &       \\
\cmidrule{2-8}
    & Positive-Qualitative        & 2     &       & 5     & 2     &       &       \\
    & Neutral-Qualitative         & 1     & 2     & 18    & 5     &       & 4     \\
    & Negative-Qualitative        & 51    & 62    & 45    & 110   & 24    & 37    \\
    & ~~Dehumanization            & 8     & 4     & 3     & 26    & 3     & 13    \\
\cmidrule{2-8}
    & Positive-Behavioral         & 4     & 5     & 10    & 5     & 7     &       \\
    & Neutral-Behavioral          & 6     & 2     & 38    & 5     & 5     & 2     \\
    & Negative-Behavioral         & 116   & 90    & 170   & 135   & 139   & 76    \\
\bottomrule
\end{tabular}
\end{table}

\subsection{Example of Data and Annotation}
To get an idea of what kinds of comments make up ToxiREX, we present a full thread in \autoref{tab:example_thread}.
We can see that the sub-thread contains implications, counterspeech, and clearly non-toxic comments.
The thread consists of comments that were left on a post titled: ``Ukrainians were among tens of thousands of LGBTQ activists who took part in the Polish capital's Pride parade Saturday, in a country hosting tens of thousands of Ukrainians who have fled the Russian invasion''.
It features two users discussing whether refugees should ``participate in a protest while they’re in a country that isnt theirs''.

The discussion starts with user A asking if Ukrainian refugees are engaging in bad behavior by participating.
User B challenges this, and uses a (somewhat extreme) hypothetical scenario to demonstrate their point.
After some back and forth, the exchange ends with both users accusing each other of bad behavior, such as ``justifying rape'', and being ``homophobic''.

Our dataset contains many examples of such political conversations. 
They often include users making controversial claims that could be interpreted as toxic, and then devolve into creative name calling.
We believe this type of data is underexplored compared to more obviously harmful text.

\subsection{Initial Findings}

\paragraph{Statistics}
Sub-thread lengths, the number of potentially toxic comments, and the frequencies of implication properties and polarities can be seen in \autoref{tab:test_set_statistics}.
The number of comments considered potentially toxic is close to or over half, depending on the language.
This is higher than levels of toxicity reported for other datasets \citep{vidgen_directions_2020}. 
There are two reasons for this. 
First, we specifically ask annotators if a comment \textit{might} be considered toxic.
And second, we used an LLM to pre-select sub-threads based on whether it contained any toxicity.

The testset contains diverse implications, although situational implications are under\-rep\-re\-sent\-ed compared to qualitative and behavioral implications.
Problematic annotations for the implication category (\texttt{implCategory}) and polarity (\texttt{implPolarity}) variables are rare. 
These include implications being labeled under the `NonSpecific' category, as well as those missing a polarity.
From these results, we can conclude that the implication properties included in the schema cover a large majority of the potentially toxic implications present in our data.
%

\paragraph{Inter-annotator Agreement}
The agreement, calculated independently for the categorical variables, can be seen in \autoref{tab:naive_agreement}. 
%
Going by \citet{landis_measurement_1977}, agreement is fair to moderate (between 0.2 and 0.6) for the \texttt{toxicity} and \texttt{justInappropriate} label.
Agreement for the variables related to the implication are often lower (ranging from -0.3 to 0.5).
However, given the hierarchical nature of the annotation task, disagreements are not immediately indicative of a problem.
As mentioned previously, annotators were instructed to annotate with respect to what they believed to be the main implication of a comment.
Because annotators could have different implications in mind, a degree of disagreement is expected for these variables.

Furthermore, different languages were annotated by different annotators, making agreement vary by language.
While Arabic, German, Spanish and Turkish were annotated by students, the agreement does not appear to be systematically lower than for English and Dutch, which were annotated by experts. 
For some languages the annotators came from more diverse backgrounds, for example for Arabic, annotators came from Morocco, Egypt, and Jordan, this could be another contributor to variation in agreement.

It is also likely that agreement would be higher if measured by sub-thread, since different annotators might differ on which comments they believe to be the toxic ones, while agreeing on the presence and type of toxicity.

\begin{table}
\small
\centering
\newcommand{\cn}[1]{\mc{1}{c}{#1}}
\caption{The agreement scores (Fleiss's Kappa), computed naively.}
\begin{tabular}{rrrrrrrr}
\toprule
     & \multicolumn{1}{l}{\textbf{AR\rule{8mm}{0pt}}} 
     & \multicolumn{1}{l}{\textbf{DE\rule{8mm}{0pt}}} 
     & \multicolumn{1}{l}{\textbf{EN\rule{8mm}{0pt}}} 
     & \multicolumn{1}{l}{\textbf{ES\rule{8mm}{0pt}}} 
     & \multicolumn{1}{l}{\textbf{NL\rule{8mm}{0pt}}} 
     & \multicolumn{1}{l}{\textbf{TR\rule{8mm}{0pt}}} 
     & \multicolumn{1}{l}{Average}  \\
\midrule
    \textbf{toxicity} & \cc{B9E3CE}0.42 & \cc{CAEADA}0.32 & \cc{C5E8D7}0.35 & \cc{A6DBC1}0.53 & \cc{CAEADA}0.32 & \cc{9BD7B9}0.60 & \cc{B9E3CE}0.42 \\
    \textbf{countnernarrative} & \cc{F2FAF6}0.08 & \cc{F2FAF6}0.08 & \cc{CAEADA}0.32 & \cc{D7EFE3}0.24 & \cc{D5EEE2}0.25 & \cc{D0ECDF}0.28 & \cc{E0F3E9}0.19 \\
\midrule
    \\
\midrule
    \multicolumn{8}{l}{\small Below is reported only on cases where there was agreement on toxicity and counternarrative.} \\
\midrule
    \textbf{SUPPORT} & 84 & 57 & 61 & 114 & 29 & 91 & \multicolumn{1}{l}{\cc{F3F3F3}} \\
    \textbf{justInappropriate} & \cc{DEF2E8}0.20 & \cc{C5E8D7}0.35 & \cc{CFECDE}0.29 & \cc{B6E2CC}0.44 & \cc{A0D9BD}0.57 & \cc{A3DABF}0.55 & \cc{BBE4D0}0.41 \\
\midrule
    \\
\midrule
    \multicolumn{8}{l}{\small Below is reported only on cases where each annotator agreed there was some toxic implication present.} \\
\midrule
    \textbf{SUPPORT} & 28 & 21 & 46 & 62 & 29 & 19 & \multicolumn{1}{l}{\cc{F3F3F3}} \\
    \textbf{subject} & \cc{EDF8F3}0.11 & \cc{C0E6D3}0.38 & \cc{ABDDC5}0.50 & \cc{BBE4D0}0.41 & \cc{77C8A1}0.81 & \cc{FAFDFC}0.03 & \cc{C3E7D5}0.36 \\
    \textbf{implPolarity} & \cc{DBF1E6}0.22 & \cc{FEFCFC}-0.02 & \cc{EDF8F3}0.11 & \cc{CFECDE}0.29 & \cc{FDF8F8}-0.05 & \cc{FEFCFC}-0.02 & \cc{E8F6EF}0.14 \\
    \textbf{implCategory} & \cc{D0ECDF}0.28 & \cc{CFECDE}0.29 & \cc{F5FBF8}0.06 & \cc{C1E6D4}0.37 & \cc{CDEBDC}0.30 & \cc{E0F3E9}0.19 & \cc{D7EFE3}0.24 \\
    \textbf{implSarcasm} & \cc{E6F5EE}0.15 & \cc{C5E8D7}0.35 & \cc{EAF7F0}0.13 & \cc{E6F5EE}0.15 & \cc{B2E0CA}0.46 & \cc{FDF5F5}-0.07 & \cc{E0F3E9}0.19 \\
    \textbf{implStereotype} & \cc{F4FBF7}0.07 & \cc{C5E8D7}0.35 & \cc{FBEBEA}-0.15 & \cc{FEFDFD}-0.01 & \cc{E3F4EC}0.17 & \cc{F6D3D0}-0.33 & \cc{FAFDFC}0.03 \\
\bottomrule
\end{tabular}
\label{tab:naive_agreement}
\end{table}

\subsection{Analysis of Annotator Disagreements}
Given that not all disagreements are problematic in our case, we seek to better understand what is behind the disagreements in our annotations.
Thus, we set out to categorize each disagreement by what we believe to be the source of the disagreement.
The disagreements were categorized as:
\begin{itemize}
    \item \textbf{error}, indicating the disagreement is due to an annotation error by one or more annotators;
    \item \textbf{orthogonal}, signifying that there is a disagreement in name only, because the other answers are also correct;
    \item \textbf{conditional}, meaning the disagreement is a direct consequence of an upstream disagreement, not a disagreement in and of itself;
    \item \textbf{subjective}, for disagreements that results from a difference in perspective; or,
    \item \textbf{disjunct}, indicating we were unable to reduce the disagreement to one of the above cases.
\end{itemize}

Because of the labor-intensive nature of this analysis, it was done only for a limited subset of the annotation. 
First, it was done only for the English part of the data.
Second, we restricted the analysis to the subset of comments for which annotators agreed on the first four annotation variables.
Last, we only categorized disagreements for those variables essential for capturing what is implied by a comment.
These variables are those (partially) inside the box labeled `Proposition-Composition' in \autoref{fig:annotation_variable_schema}.
This leaves 47 out of the 499 English comments in the pre-selection, coming from 38 out of 102 sub-threads.

The first stage of this analysis consisted of the annotators going over their annotations again, to correct any errors they observed with the benefit of both hindsight and access to the other annotators' choices. 
Simultaneously, while checking for errors, the annotators indicated for each disagreement which of the categories given above was most applicable. 

In the second stage, the variable-level categorizations of each annotator are considered to decide which category applies best for each variable of each comment. 
This was partially done by majority vote, and partially based on discussions between the annotators.

We noticed that disagreements between individual variables often cannot be understood independently, often making it difficult to categorize the disagreement. 
So, in the third stage, we supplemented the analysis with proposition-level categorizations.
Taking the corrected errors into account, we generated the summary sentences again. 
Then, for each comment, the variable-level categorizations of disagreements are considered alongside the summary sentences to come to a propo\-si\-tion-level categorization.
When the propo\-si\-tion-level process gave additional insight into the cause of the disagreement, this was also used to adjust categorizations on the variable-level categorizations.

\begin{table}
\small
\let\mc\multicolumn
\let\rot\rotatebox
\newcommand{\cn}[1]{\mc{1}{c}{#1}}
\let\hp\hphantom
\centering
\caption{The statistics of the categories that were found to be most descriptive of why a given disagreement between annotators occurred. Disagreement categories are in bold, aggregates are italicized. Rows where the categories are accompanied by a `+' show the frequency of that category occurring in combination with one or more of the categories that occur above it in the table. For example, the row labeled subjective+ presents frequencies of subjective occurring together with orthogonal and/or error. The first row marked `\#' shows the number of comments that were not marked as `conditional'. The statistics are computed over the comments not marked `conditional'.}%
\label{tab:analysis_statistics}
\begin{tabular}{@{}lrrrrrrrrrr@{}}
\toprule
 &
  \mc{1}{l}{\rot{60}{subject}} &
  \mc{1}{l}{\rot{60}{subjectGroup}} &
  \mc{1}{l}{\rot{60}{hasOther}} &
  \mc{1}{l}{\rot{60}{other}} &
  \mc{1}{l}{\rot{60}{otherGroup}} &
  \mc{1}{l}{\rot{60}{implPolarity}} &
  \mc{1}{l}{\rot{60}{implCategory}} &
  \mc{1}{l}{\rot{60}{implTemporality}} &
  \mc{1}{l}{\rot{60}{\textit{proposition}}} \\
\cmidrule(){1-1} \cmidrule(l){2-3} \cmidrule(l){4-6} \cmidrule(l){7-9} \cmidrule(l){10-10} 
\#                          & 47             & 46             & 47             & 22             & 31             & 47             & 47             & 47             & 47   \\ 
\cmidrule(){1-1} \cmidrule(l){2-3} \cmidrule(l){4-6} \cmidrule(l){7-9} \cmidrule(l){10-10} 
{agreement}                 & 68\%           & 47\%           & 36\%           & 82\%           & 65\%           & 55\%           & 38\%           & 34\%           &  4\%  \\  
\cmidrule(){1-1} \cmidrule(l){2-3} \cmidrule(l){4-6} \cmidrule(l){7-9} \cmidrule(l){10-10} 
\textbf{error}              &  6\%           &  2\%           & 13\%           & 18\%           & 10\%           & 21\%           & 11\%           &  4\%           &  6\%  \\
\textit{sum}                & \textit{74\%}  & \textit{49\%}  & \textit{49\%}  & \textit{100\%} & \textit{74\%}  & \textit{77\%}  & \textit{49\%}  & \textit{38\%}  & \textit{11\%} \\
\cmidrule(){1-1} \cmidrule(l){2-3} \cmidrule(l){4-6} \cmidrule(l){7-9} \cmidrule(l){10-10} 
\textbf{orthogonal}         & 11\%           & 28\%           & 23\%           &  0\%           &  3\%           &  0\%           & 23\%           & 34\%           & 17\%  \\
\textbf{orthogonal}+        & \textit{0\%}   & \textit{2\%}   & \textit{0\%}   & \textit{0\%}   & \textit{0\%}   & \textit{2\%}   & \textit{0\%}   & \textit{4\%}   & \textit{15\%} \\ 
\textit{sum}                & \textit{85\%}  & \textit{79\%}  & \textit{72\%}  & \textit{100\%} & \textit{77\%}  & \textit{79\%}  & \textit{72\%}  & \textit{77\%}  & \textit{43\%} \\ 
\cmidrule(){1-1} \cmidrule(l){2-3} \cmidrule(l){4-6} \cmidrule(l){7-9} \cmidrule(l){10-10} 
\textbf{subjective}         & 13\%           & 15\%           & 17\%           &  0\%           &  6\%           &  4\%           & 15\%           & 19\%           & 21\%  \\
\textbf{subjective}+        &  0\%           &  2\%           &  2\%           &  0\%           &  13\%          &  2\%           & 0\%            & 0\%            & 15\%  \\
\textit{sum}                & \textit{98\%}  & \textit{96\%}  & \textit{91\%}  & \textit{100\%} & \textit{97\%}& \textit{85\%}& \textit{87\%}& \textit{96\%}& \textit{79\%} \\ 
\cmidrule(){1-1} \cmidrule(l){2-3} \cmidrule(l){4-6} \cmidrule(l){7-9} \cmidrule(l){10-10} 
\textbf{disjunct}           &  2\%           &  4\%           &  9\%           &  0\%           &  0\%           &  9\%           &  9\%           &  4\%           &  9\%  \\
\textbf{disjunct}+          &  0\%           &  0\%           &  0\%           &  0\%           &  3\%           &  6\%           &  4\%           &  0\%           &  13\%  \\
\textit{sum}                & \textit{100\%} & \textit{100\%} & \textit{100\%} & \textit{100\%} & \textit{100\%} & \textit{100\%} & \textit{100\%} & \textit{100\%} & \textit{100\%} \\
\bottomrule
\end{tabular}
\end{table}

\begin{table}%
\small%
\centering%
\caption{Examples of each of the types of disagreement. The answers given by annotators are represented by the \textit{Summary Sentences}. Newline characters in the the comments were replaced by the `\ensuremath{\hookleftarrow}' character.}%
\label{tab:disagreement_examples}%
\vspace{-5mm}%
\begin{subtable}{\textwidth}
\subcaption{Orthogonal.}%
\begin{tabular}{@{}p{7.2cm}lp{4.6cm}@{}}
\toprule
    \textbf{Comment} & \textbf{Category} & \textbf{Summary Sentences} \\ 
\midrule
    \multirow{5}{=}{The rise of Islam becoming a state religion in Turkey is the reason why I think the "God is punishing them' comments are going on. Edrogan has abandoned secularism to try to turn Turkey into an Islamic state and now this earthquake. Is it God punishing them? I HIGHLY doubt it (agnostic here) but hopefully the Turkish people think it is so they don't descend into a full religious dictatorship.} 
    & \multirow{6}{*}{orthogonal} 
       & \textbullet~Muslim Turks had/have harm come to them.   \\[2mm]
    &  & \textbullet~An individual [President Erdogan] has negative behavior. \\[2mm]
    &  & \textbullet~Erdogan government had negative behavior. \\[2mm]
\midrule
    \multirow{8}{=}{All jokes aside.. This protest showed most controversial police brutality we've seen in recent years in Belgium, with multiple claims of unlawful enforcement including but not limited to: \ensuremath{\hookleftarrow} \ensuremath{\hookleftarrow} Unprovoked use of the water cannon.\ensuremath{\hookleftarrow} Unprovoked use of teargas canisters on protestors (including woman and children within the crowd) during a speech.\ensuremath{\hookleftarrow} Police refusing to grand medical aid to wounded unconscious protester, who was dragged by fellow protestors towards them, refusing said protestor's passage past a barbed wire barricade.\ensuremath{\hookleftarrow} \ensuremath{\hookleftarrow} Footage of this all over the place.. yet there was a serious lack of media coverage on the entire event, which was expecting 10k visitors but ended up around 35k.\ensuremath{\hookleftarrow} \ensuremath{\hookleftarrow} Not looking to start an pro/anti argument, simply providing some background to the event. } 
    & \multirow{12}{*}{orthogonal} 
       & \textbullet~the Belgian police had negative behavior compared to protesters.  \\ \\ \\ 
    &  & \textbullet~Belgian police has negative behavior.                             \\ \\ \\ 
    &  & \textbullet~Belgian police had negative behavior.                             \\ \\ \\[1mm] 
\bottomrule
\end{tabular}
\end{subtable}
\end{table}

\begin{table}
\ContinuedFloat
\centering
\small
\renewcommand{\arraystretch}{1.3}
\begin{subtable}{\textwidth}
\subcaption{Subjective \& disjunct.}
\label{tab:disagreement_examples_B}
\begin{tabular}{@{}p{6.5cm}lp{5.4cm}@{}}
\toprule
    \textbf{Comment} & \textbf{Category} & \textbf{Summary Sentences} \\ 
\midrule
    \multirow{3}{=}{Yes, it's not a legal case, it's a political decision. It doesn't matter if all 10k rioters say that Trump incited them.  We already know he provably did and if this were a real jury, he'd be convicted. We need to put the stains on the GOP's "blue lives matter" dress force them to bleach themselves.} 
    & \multirow{4.5}{*}{subjective} 
       & \textbullet~far right Republicans supporting Trump had/have/will have negative behavior.  \\
    &  & \textbullet~The Republican Party (GOP) has negative behavior.  \\
    &  & \textbullet~Republicans had negative behavior.  \\
\midrule
    \multirow{3}{=}{But they're really bad at it. They just lie and immediately get mocked for it. It's just a terrible look for Israel. If you can't be honest how will anyone take you seriously or get people to be sympathetic to you stealing land.} 
    & \multirow{4.5}{*}{subjective} 
       & \textbullet~people supporting Israel including the people of Israel themselves have/will have negative behavior.  \\
    &  & \textbullet~Israel has negative behavior.  \\
    &  & \textbullet~Israel had negative behavior.  \\
\midrule
    \multirow{3}{=}{My argument? \ensuremath{\hookleftarrow} \ensuremath{\hookleftarrow} Antifa = proud boys, put them in the same jail cell and let them fight it out there. \ensuremath{\hookleftarrow} \ensuremath{\hookleftarrow} r/law: dOwNvOTe!} 
    & \multirow{5}{*}{disjunct} 
       & \textbullet~Antifa has harm come to them compared to proud boys. \\
    &  & \textbullet~Antifa and proud boys will have negative behavior.  \\
    &  & \textbullet~Antifa and the proud boys will have harm come to them.  \\
\midrule
    \multirow{3}{=}{And we all know what the Russians are gonna do:\ensuremath{\hookleftarrow} \ensuremath{\hookleftarrow} Go to Moldova and kill the Russians first. Then forcibly conscript the rest of the men and rape the Russian women remaining.} 
    & \multirow{3}{*}{disjunct} 
       & \textbullet~Russians will have negative nature.   \\
    &  & \textbullet~Russians will have negative behavior. \\
    &  & \textbullet~Russia will have negative behavior.   \\ 
\bottomrule
\end{tabular}
\end{subtable}
\end{table}

\paragraph{Findings}
The statistics of how many times each category was chosen can be seen in \autoref{tab:analysis_statistics}.
A selection of examples is given in \autoref{tab:disagreement_examples}.

When annotators categorize an apparent disagreement as {orthogonal}, they do not truly disagree with the other annotators.
While they chose to annotate differently, they acknowledge that the other annotations are equally and simultaneously correct.
The percentage of comments where the annotators chose the exact same answers before error correction ranges from 34\% to 82\%.
When including the comments marked {orthogonal} and the error correction, the percentage is at least 72\% for all variables.
This shows that, as expected, a large portion of apparent disagreements, were not truly disagreements at all.

Looking at the remaining disagreements, most are {subjective}.
When the annotators marked something {subjective}, it is usually due to a different interpretation of the comment.
For these cases, the annotators do disagree with each other, but still recognize that the other annotations rest on a valid understanding of the comment and annotation schema.

Only a small fraction of the disagreements did not fall into one of the other categories (ranging from 0\% to 9\%), and were thus marked disjunct.
These are usually cases where the annotators fundamentally disagreed with another annotator's interpretation.
Given these results, we believe the data to be of high quality.

\section{Training Data Annotation}
To produce silver-standard annotations for the training data, we use OpenAI's GPT4o model. 
The model is used with the `Structured Outputs' feature which uses the pydantic model described below in \ref{ssec:structured_outputs}.
Sub-threads are presented to the model as shown in \autoref{prompt:silver_annotation}. 
If a sub-thread has no post text, it is instantiated with the value ``EMPTY''.

\subsection{Structured Outputs for Toxic Reasoning} \label{ssec:structured_outputs}
To make an LLM produce outputs in accordance with the toxic reasoning schema, we encode a version of the schema as a pydantic\footnote{https://docs.pydantic.dev/} model.

The pydantic model is converted into a JSON Schema\punctfootnote[.][0.5]{https://json-schema.org/}. 
The role of the JSON schema is two-fold. 
First, with this JSON schema, models can be forced to produce responses which are valid instances of the toxic reasoning schema. 
To make this work, a sampling strategy is used that masks tokens which would produce invalid outputs \citep{willard_efficient_2023}. 
Second, the JSON schema is provided in the input to the model.
For an appropriately trained model, this largely replaces the role of the instructions provided to a model. 
As such, the pydantic model also includes textual descriptions of each field, which guide the LLM on how to fill out those fields.

In total, the pydantic model consists of 9 Python classes.
The main class is the \texttt{ToxicReasoning} class, every sub-thread is provided to the LLM as one input, and the model is asked to produce one \texttt{ToxicReasoning} instance per comment that is deemed toxic.
The full pydantic code can be found in \autoref{apx:pydantic_trs1}.

\begin{prompt}
\begin{lstlisting}[style=llmprompt]
From a thread in r/{subreddit}

Post Title: `{title}`
Post Text: ```
{text}
```

Message {i} by {author_name} on {creation_time}:
```
{comment_body}
```

Please indicate for each of the messages in this thread whether they include toxic language, using the specified schema.
\end{lstlisting}%
\vspace{-2mm}%
\caption{Template string containing messages in a sub-thread and instruction. Lines 8 through 12 are repeated and instantiated independently for each message.}
\label{prompt:silver_annotation}
\end{prompt}

\subsection{Statistics}
Statistics of the training set can be seen in \autoref{tab:train_set_statistics}.
Roughly 30\%-40\% of comments were considered potentially toxic by GPT4o, varying by language.
This is lower than the amount considered potentially toxic by human annotators (closer to 50\%).
Compared to human annotators, GPT4o is also somewhat more likely to annotate negative polarity and qualitative category implications.

Note that the test set data was selected based on a stratified sample, but the train set was not. 
Thus, differences in these statistics cannot be attributed purely to differences between GPT4o and the human annotators.


\begin{table}
\small
\centering
\let\mc\multicolumn
\newcommand{\dd}{$^\dagger$}
\caption{Train set statistics, including label statistics. The first three rows give sub-thread lengths. The rest of the rows show the number of annotations which has that category or polarity.}
\label{tab:train_set_statistics}
\begin{tabular}{@{}l@{\hspace{1mm}}lrrrrrr@{}}
\toprule
 & & \mc{1}{l}{\textbf{AR}} & \mc{1}{l}{\textbf{DE}} & \mc{1}{l}{\textbf{EN}} & \mc{1}{l}{\textbf{ES}} & \mc{1}{l}{\textbf{NL}} & \mc{1}{l}{\textbf{TR}} \\
    & Nr. sub-threads             & 296   & 1,421 & 17,833& 3,340 & 612   & 940   \\
\midrule
    & Avg. sub-thread length       & 5.51  & 5.16  & 5.20  & 5.21  & 5.39  & 5.43  \\
    & Min. sub-thread length       & 4     & 4     & 4     & 4     & 4     & 4     \\
    & Max. sub-thread length       & 10    & 10    & 10    & 10    & 10    & 10    \\
\midrule
    & Nr. of non-toxic            & 1,034 & 5,389 & 62,224& 10,533& 2,249 & 3,097 \\
    & Nr. of possibly toxic       & 597   & 1,944 & 30,493& 6,858 & 1,049 & 2,005 \\
\midrule
\multirow{7}{*}{\rotatebox{90}{\textbf{Category}}}
    & Situational                 & 33  & 149   & 1,950 & 477   & 51    & 81  \\
    & ~~Harm                      & 28	& 176	& 2,957 & 571   & 97    & 148 \\
    & Qualitative                 & 223	& 583	& 8,705	& 2,496	& 369	&603  \\
    & ~~Dehumanization            & 23	& 83	& 763	& 333	& 25	&148  \\
    & Behavioral                  & 240	& 799	& 13,436& 2,412	& 423	&574  \\
    & NonSpecific                 & 19	& 80	& 1,623	& 230	& 39	&36   \\
    & UnclearOrNone               & 0	& 1	    & 17	& 3	    & 3	    &1    \\
\midrule
\multirow{3}{*}{\rotatebox{90}{\textbf{Polarity}}}
    & Positive                    & 9	& 6	    & 102	& 20	& 23    &      \\
    & Neutral                     & 2	& 16	& 189	& 51	& 12	& 10   \\
    & Negative                    & 555	& 1,849	& 29,160& 6,451	& 993	& 1,578\\
\bottomrule
\\
\toprule
\multirow{11}{*}{\rotatebox{90}{\textbf{Polarity~+~Category}}}
    & Positive-Situational        & 1   &       & 7     & 1     &       &     \\
    & Neutral-Situational         &     & 4     & 13    & 5     &       &     \\
    & Negative-Situational        & 32  & 145   & 1,930 & 471   & 51    & 77  \\
    & ~~Harm                      & 28  & 176   & 2,957 & 571   & 97    & 148 \\
\cmidrule{2-8}
    & Positive-Qualitative        &     &       & 12    & 2     &       &     \\
    & Neutral-Qualitative         &     & 1     & 26    & 13    & 2     & 2   \\
    & Negative-Qualitative        & 223 & 582   & 8,667 & 2,481 & 367   & 601 \\
    & ~~Dehumanization            & 23	& 83	& 763	& 333	& 25	& 148 \\
\cmidrule{2-8}
    & Positive-Behavioral         & 7   & 6     & 53    & 14    & 1     &     \\
    & Neutral-Behavioral          & 2   & 8     & 87    & 22    & 7     & 1   \\
    & Negative-Behavioral         & 231 & 785   & 13,296& 2,376 & 415   & 573 \\
\bottomrule
\end{tabular}
\end{table}

\section{Baseline Experiments}
In this section, we provide baseline results on our dataset.
We evaluate OpenAI's GPT4o, a commercially available LLM, in a zero-shot setup. 
We also fine-tune XLM-RoBERTa \citep{conneau_unsupervised_2020} on the training set using a simple baseline setup.

\subsection{A Fine-tuning Baseline}\label{ssec:fine-tuning_approach}
The main approach we evaluate works by adding classification heads to a pre-trained language model.
We use this strategy in combination with XLM-RoBERTa-large \citep{conneau_unsupervised_2020}, and EuroBERT \citep{boizard_eurobert_2025}. 
A sub-thread is provided to the model as input, the same way as shown in \autoref{prompt:silver_annotation}. 
However, one change is made: 
we add a special `\texttt{<COMMENT>}' token to the vocabulary and randomly initialize its embedding. 
Then, after each comment in the sub-thread we add this token.
It is the representations for these `\texttt{<COMMENT>}' tokens that we use as input to the classification heads.
The only exception are the \texttt{subjectTokens}, \texttt{otherTokens}, and \texttt{implTokens} variables, which are predicted using binary classification heads on the tokens of each comment. 

This approach is a relatively straightforward way of getting a model to make predictions according to the toxic reasoning schema. 
Unlike GPT4o, which predicts each field in sequence, this model outputs predictions on all variables `simultaneously'. 
Given two variables A and B, where A precedes B in the hierarchy, this setup does not allow a model to adjust its prediction of variable B to the its prediction on variable A.

\begin{figure}
    \centering
    \includegraphics[width=.85\linewidth]{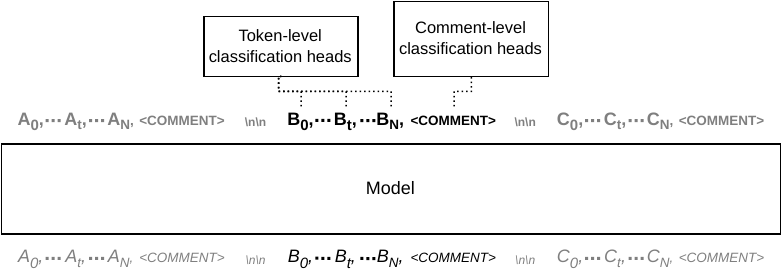}
    \caption{Schematic view of approach for fine-tuning baseline. From bottom to top, we see: input tokens for comments A, B and C; the model; output hidden states; and finally, the classification heads.}
\end{figure}

\paragraph{Alternative Strategy}
We also explored fine-tuning a model for producing instances of the pydantic model in the same way that GPT4o does. 
Preliminary experiments using this strategy did not result in better performance than the strategy given above (see \autoref{apx:decoder_strategy}). 
Since our primary goal was to produce a simple baseline, we leave such strategies for future work.

\subsection{Evaluation}
In our setup, every annotator potentially has a different implication in mind when producing their annotations. 
This presents two problems for the use of ordinary classification metrics, like F1 scores.

\paragraph{Problem 1}
For a given variable $V$, comparing model predictions against annotations only makes sense if the annotation and prediction agree on the variables which precede $V$ in the hierarchy of \autoref{fig:annotation_variable_schema}.
Disagreements on earlier variables can cause predictions and annotations to disagree on whether there is an implication at all. 
Thus, the model may produce a value for a given variable, while the annotator did not, or vice versa. 
To address this, we consider the following variants of the metrics.
\begin{itemize}
    \item \textbf{propagated} --- If a model does not produce any implication when it should have, or does when it should not have, then propagate this error to all subsequent variables as well.
    \item \textbf{conditional} --- Report the metric on the subset of comments/annotations for which both the model and annotator agreed that the variable should be annotated (i.e. that none of the preceding variables had a value that should terminate the prediction/annotation). 
    \item \textbf{optimistic} --- Models like that described in \autoref{ssec:fine-tuning_approach} can produce predictions for all variables even if the prediction for an earlier variable would require ignoring later predictions (e.g. \texttt{hasImplication} being false). This allows us to evaluate later predictions as though the prediction for the earlier variable was correct, even if it was not.
\end{itemize}

\paragraph{Problem 2}
Combining annotations to produce a single label, often done by majority vote, similarly requires that annotations agree on the variables which precede it in the hierachy.
Rather than combine the annotations, we compare predictions against individual annotations. 
Various strategies have been used to evaluate on datasets that contain multiple annotations per item \citep{uma_learning_2021}.
We use three approaches that allow the use of ordinary classification metrics:
\begin{itemize}
    \item \textbf{all}, do not select one annotation per comment, but repeat comments such that each annotation is included independently; 
    \item \textbf{random}, sample one annotation per comment, repeating N times, then report the median; and,
    \item \textbf{max}, choose the annotation for which another (set of) variable is most similar to the prediction.
\end{itemize}

\paragraph{Metrics}
For the evaluation, we report the average precision, recall, and F1, with each being calculated per class first.
The number we report is the weighted average over the classes, where the weights are proportional to the frequency of the class.

We compare these scores against what we call the `Prior' baseline. 
This is the score obtained by guessing classes proportional to their frequency in the label distribution of the test set.

\begin{table}
\small
\centering
\caption{Weighted F1, recall (Rec.), and precision (Prec.) scores, as well as the  `Prior' baseline, calculated per language. Left column shows English, and the right shows average over all languages. Labels obtained by majority vote amongst annotators. F1 scores that do not beat the `Prior' baseline are underlined. Darker shades of blue correspond to higher scores.}
\label{tab:f1_majority}
\begin{tabular}{@{}l@{\hspace{1mm}}llrrrrrrrr}
\toprule
 & & & \multicolumn{4}{c}{English} & \multicolumn{4}{c}{All} \\ 
\cmidrule(lr){4-7} \cmidrule(lr){8-11} 
 & & & \multicolumn{1}{c}{F1}  & \multicolumn{1}{c}{Prec.}  & \multicolumn{1}{c}{Rec.}  & \multicolumn{1}{c}{Prior}  
     & \multicolumn{1}{c}{F1}  & \multicolumn{1}{c}{Prec.} & \multicolumn{1}{c}{Rec.}  & \multicolumn{1}{c}{Prior} \\ 
\midrule
\multirow{5}{*}{\rotatebox{90}{GPT4o}}  &
\multirow{5}{*}{\rotatebox{90}{(prop.)}} 
  & toxicity          &\cc{C0D5F8}    .74  &\cc{BFD4F8}.75 &\cc{BFD4F8}.75  &\cc{DBE7FB}.52 &\cc{BDD3F7}    .77 &\cc{BAD1F7}.79 &\cc{BDD3F7}.77 &\cc{DEE9FB}.50 \\
 && counternarrative  &\cc{A7C4F5}    .95  &\cc{A8C5F5}.94 &\cc{A6C3F5}.96  &\cc{A8C5F5}.94 &\cc{AFCAF6}    .88 &\cc{B4CDF6}.84 &\cc{ABC7F5}.92 &\cc{B3CCF6}.85 \\
 && justInappropriate &\cc{B1CBF6}    .87  &\cc{B2CBF6}.86 &\cc{ADC8F6}.90  &\cc{B5CEF7}.83 &\cc{BCD2F7}    .78 &\cc{B7CFF7}.82 &\cc{B5CEF7}.83 &\cc{C4D8F8}.71 \\
 && hasImplication    &\cc{BCD2F7}    .78  &\cc{BAD1F7}.79 &\cc{BCD2F7}.78  &\cc{D1E1FA}.60 &\cc{BED4F8}    .76 &\cc{B9D0F7}.80 &\cc{C0D5F8}.74 &\cc{C8DAF9}.68 \\
 && hasOther          &\cc{ECF2FD}{\ul.38} &\cc{BDD3F7}.77 &\cc{F8FBFF}.28 &\cc{CBDDF9}.65 &\cc{E6EEFC}{\ul.43}&\cc{B9D0F7} .80 &\cc{F3F7FE}.32 &\cc{C3D7F8}.72 \\
\midrule
\multirow{5}{*}{\rotatebox{90}{XLM-R}} &
\multirow{5}{*}{\rotatebox{90}{(prop.)}} 
  & toxicity          & \cc{C9DBF9}    .67 &\cc{C6D9F9}.69 &\cc{C6D9F9}.69 &\cc{DBE7FB}.52 &\cc{C9DBF9}    .67 &\cc{C3D7F8}.72 &\cc{C6D9F9}    .69 &\cc{DEE9FB}.50 \\
 && counternarrative  & \cc{A7C4F5}    .95 &\cc{A8C5F5}.94 &\cc{A4C2F4}.97 &\cc{A8C5F5}.94 &\cc{B1CBF6}    .87 &\cc{B4CDF6}.84 &\cc{ACC7F5}    .91 &\cc{B3CCF6}.85 \\
 && justInappropriate & \cc{B1CBF6}    .87 &\cc{ABC7F5}.92 &\cc{ACC7F5}.91 &\cc{B5CEF7}.83 &\cc{BED4F8}    .76 &\cc{B3CCF6}.85 &\cc{B7CFF7}    .82 &\cc{C4D8F8}.71 \\
 && hasImplication    & \cc{BFD4F8}    .75 &\cc{BFD4F8}.75 &\cc{BED4F8}.76 &\cc{D1E1FA}.60 &\cc{BDD3F7}    .77 &\cc{BDD3F7}.77 &\cc{BED4F8}    .76 &\cc{C8DAF9}.68 \\
 && hasOther          & \cc{F2F7FE}{\ul.33}&\cc{C5D8F8}.70 &\cc{FFFFFF}.22&\cc{CBDDF9}.65 &\cc{F1F6FE}{\ul.34}&\cc{BCD2F7}.78 &\cc{FFFFFF}.22 &\cc{C3D7F8}.72 \\ 
\midrule
\multirow{5}{*}{\rotatebox{90}{XLM-R}} &
\multirow{5}{*}{\rotatebox{90}{(opt.)}}
 & toxicity           & \cc{C9DBF9}.67 & \cc{C6D9F9}.69 & \cc{C6D9F9}.69 & \cc{DBE7FB}.52 & \cc{C9DBF9}.67 & \cc{C3D7F8}.72 & \cc{C6D9F9}.69 & \cc{DEE9FB}.50 \\
 && counternarrative  & \cc{A7C4F5}.95 & \cc{A8C5F5}.94 & \cc{A4C2F4}.97 & \cc{A8C5F5}.94 & \cc{B1CBF6}.87 & \cc{B4CDF6}.84 & \cc{ACC7F5}.91 & \cc{B3CCF6}.85 \\
 && justInappropriate & \cc{B1CBF6}.87 & \cc{ABC7F5}.92 & \cc{ACC7F5}.91 & \cc{B5CEF7}.83 & \cc{BED4F8}.76 & \cc{B3CCF6}.85 & \cc{B7CFF7}.82 & \cc{C4D8F8}.71 \\
 && hasImplication    & \cc{BFD4F8}.75 & \cc{BFD4F8}.75 & \cc{BED4F8}.76 & \cc{D1E1FA}.60 & \cc{BDD3F7}.77 & \cc{BDD3F7}.77 & \cc{BED4F8}.76 & \cc{C8DAF9}.68 \\
 && hasOther          & \cc{C6D9F9}.69 & \cc{C2D6F8}.73 & \cc{C9DBF9}.67 & \cc{CBDDF9}.65 & \cc{C4D8F8}{\ul.71} & \cc{BDD3F7}.77 & \cc{C9DBF9}.67 & \cc{C3D7F8}.72 \\
\bottomrule
\end{tabular}
\end{table}

\begin{table}
\small
\centering
\caption{The Jaccard index between GPT4o and human annotations for the token-level variables.}
\label{tab:jaccard}
\begin{tabular}{@{}lllllll@{}}
\toprule
 & \multicolumn{2}{c}{implTokens} & \multicolumn{2}{c}{otherTokens} & \multicolumn{2}{c}{subjectTokens} \\ 
\cmidrule(l){2-3} 
\cmidrule(l){4-5} 
\cmidrule(l){6-7} 
 & Jaccard & Supp. & Jaccard & Supp. & Jaccard & Supp. \\ 
\midrule
AR & .26 & 161 & .11 & 17 & .23 & 147 \\
DE & .22 & 163 & .23 & 54 & .19 & 144 \\
EN & .18 & 263 & .13 & 91 & .19 & 246 \\
ES & .25 & 280 & .20 & 45 & .20 & 249 \\
NL & .19 & 172 & .09 & 37 & .23 & 149 \\
TR & .29 & 117 & .19 & 26 & .23 & 116 \\
\bottomrule
\end{tabular}
\end{table}

\begin{table}
\small
\centering
\let\mc\multicolumn
\caption{F1 scores compared to prior baseline for PropositionComposition variables, calculated across all languages.
Evaluated in 3 scenarios, propagated, conditional on the model predicting there is an implication, and conditional on the relevant \_Tokens variable. }
\label{tab:f1_prop_comp}
\begin{tabular}{@{}llrrrrrr@{}}
\toprule
 &  & \multicolumn{3}{c}{GPT4o} & \multicolumn{3}{c}{XLM-RoBERTa} \\
\cmidrule{3-5} \cmidrule{6-8}
 &  & \mc{1}{c}{\rule{4pt}{0pt}F1\rule{4pt}{0pt}} 
                                    & \mc{1}{c}{Prior} & \mc{1}{c}{Supp.} 
                                                                & \mc{1}{c}{\rule{4pt}{0pt}F1\rule{4pt}{0pt}} 
                                                                                 & \mc{1}{c}{Prior} & \mc{1}{c}{Supp.}\\
\midrule
\multirow{6}{4mm}{\rotatebox{90}{propagated}}
& subject          & \cc{CEDFFA}.54 & \cc{D3E2FA}.49 & 1051     & \cc{DCE8FB}.39 & \cc{D3E2FA}.49 & 1051  \\
& subjectGroupType & \cc{E5EEFC}.29 & \cc{F5F8FE}.12 & 1070     & \cc{F4F8FE}.13 & \cc{F5F8FE}.12 & 1070  \\
& other            & \cc{E2ECFC}.32 & \cc{C7DAF9}.62 & 301      & \cc{E7EFFD}.27 & \cc{C7DAF9}.62 & 301   \\
& implPolarity     & \cc{C8DAF9}.61 & \cc{B8CFF7}.79 & 1169     & \cc{D7E5FB}.44 & \cc{B8CFF7}.79 & 1169  \\
& implTemporality  & \cc{CDDEF9}.55 & \cc{CDDEF9}.55 & 1180     & \cc{DCE8FB}.39 & \cc{CDDEF9}.55 & 1180  \\
& implCategory     & \cc{E1EBFC}.34 & \cc{DDE8FB}.38 & 1170     & \cc{EAF1FD}.24 & \cc{DDE8FB}.38 & 1170  \\ 
\midrule
\multirow{6}{4mm}{\rotatebox{90}{cond-impl}}
& subject          & \cc{BAD1F7}.76 & \cc{D2E1FA}.50 & 583      & \cc{BCD2F7}.74 & \cc{D2E1FA}.50 & 372   \\
& subjectGroupType & \cc{DAE6FB}.41 & \cc{F5F8FE}.12 & 636      & \cc{E8F0FD}.26 & \cc{F5F8FE}.12 & 419   \\
& other            & \cc{D4E2FA}.48 & \cc{CBDCF9}.58 & 171      & \cc{BCD2F7}.74 & \cc{C4D8F8}.65 & 68    \\
& implPolarity     & \cc{B0CAF6}.87 & \cc{B3CCF6}.84 & 627      & \cc{AFC9F6}.89 & \cc{AFCAF6}.88 & 397   \\
& implTemporality  & \cc{B8CFF7}.79 & \cc{CDDDF9}.56 & 652      & \cc{B9D1F7}.77 & \cc{CDDDF9}.56 & 419   \\
& implCategory     & \cc{D3E2FA}.49 & \cc{DFEAFC}.36 & 628      & \cc{D4E2FA}.48 & \cc{DEE9FB}.37 & 399   \\
\midrule
\multirow{6}{4mm}{\rotatebox{90}{cond-tokens}}
& subject          & \cc{B0CAF6}.87 & \cc{CCDDF9}.57 & 286      & \cc{B4CDF6}.83 & \cc{CDDDF9}.56 & 15    \\
& subjectGroupType & \cc{D4E2FA}.48 & \cc{F5F9FE}.11 & 297      & \cc{E0EAFC}.35 & \cc{F3F7FE}.14 & 17    \\
& other            & \cc{CBDCF9}.58 & \cc{CADCF9}.59 & 46       & -              & -              & 0     \\
& implPolarity     & \cc{B2CCF6}.85 & \cc{B6CEF7}.81 & 340      & \cc{BBD2F7}.75 & \cc{BED4F8}.72 & 70    \\
& implTemporality  & \cc{B3CCF6}.84 & \cc{C9DBF9}.60 & 311      & \cc{B2CCF6}.85 & \cc{CCDDF9}.57 & 62    \\
& implCategory     & \cc{D2E1FA}.50 & \cc{DEE9FB}.37 & 341      & \cc{CFDFFA}.53 & \cc{DFEAFC}.36 & 70    \\ 
\bottomrule
\end{tabular}
\end{table}

\section{Results}

We report F1, Precision, and Recall against the `Prior' baseline for the binary, multiclass and multilabel variables.
There are two exceptions.
First, for the \_Tokens variables we report the Jaccard index\punctfootnote[,][1]{The Jaccard index is defined between two sets as the size of their intersection divided by the size of their union.} because there is no fixed set of labels, as the tokens are different for each comment. 
Second, for the attitudes we do not report classification scores because the values are from an ordinal scale.
%
To assess the quality of predictions on the level of implications, including the attitudes, we also report another statistic. 
Specifically, we convert the predictions to summary sentences, as described previously, and then count how often these sentences match at least one of the human annotations.

The scores obtained by GPT4o and XLM-RoBERTa can be seen in Tables \ref{tab:f1_majority}, \ref{tab:jaccard}, \ref{tab:f1_prop_comp} and \ref{tab:implication_level_f1}.
We omit the scores obtained by EuroBERT, since they consistently underperformed XLM-R by a small margin.
In \autoref{tab:f1_majority}, we can see the classification scores obtained by GPT4o and RoBERTa for the variables we evaluate against a majority vote amongst the annotations. 
For token-level variables, \autoref{tab:jaccard} shows the Jaccard index between GPT4o and human annotations. 
In \autoref{tab:f1_prop_comp}, we can see F1 scores for variables which together describe the proposition at the heart of the implication being annotated.

In general, we found that the choice between \textbf{all}, \textbf{random}, and \textbf{max} was of little impact. 
This may be because a large share of comments were only found to have potentially toxic implications by one or two annotators (see \autoref{tab:nr_of_continuing_annotators}).
Thus, we only report \textbf{random} for all variables.

\subsection{GPT4o and XLM-R Performance}
Starting with \autoref{tab:f1_majority}, we can see that both models beat the Prior baseline for the first four variables.
With GPT4o being evaluated in zero-shot setting, it has had no exposure at all to instances of our dataset, making this performance quite impressive. 
Simultaneously, there is still room for improvement, with F1 scores for toxicity (.74), justInappropriate (.87), and hasImplication (.78) all remaining well below a perfect score.
Performance for GPT4o leads on the {toxicity} variable, but is otherwise very similar to XLM-R.
%
The performance for the \texttt{hasOther} variable is poorer, primarily due to poor recall.
For XLM-R, we also report an optimistic evaluation, where errors in the first four variables are not propagated. 
In that setting, we can see that performance for \texttt{hasOther} gets much closer to the Prior baseline.
GPT4o is not evaluated in the \textbf{optimistic} variant, because similar to the human annotation, the model only produces a full prediction if it did not encounter one of stopping conditions in the first few variables.

Looking at \autoref{tab:f1_prop_comp}, we can see that in the \textbf{propagated} setting, F1 scores for both models often do not outperform the Prior baseline.
However, when looking at \textbf{cond-impl}, which includes only the instances for which the model predicted an implication, 
we see that F1 scores are much higher.
Of course, just because the model predicted an implication, does not mean that this is the same implication the human annotator had in mind. 
Thus, we also include \textbf{cond-tokens}.
In that evaluation setting, we only include instances where the tokens selected by the model and the annotation have a Jaccard score of at least $0.5$.
We use the most relevant \texttt{\_Tokens} variable, that is: 
\texttt{subjectTokens} for the variables related to the subject, 
\texttt{otherTokens} for \texttt{other}, and 
\texttt{implTokens} for the variables related to the implication.
In this setting, we see the performance of both models rise further.
This supports the idea that the regular classification scores are too pessimistic.
At the same time, it should be noted that by selecting samples where the model agrees with human annotators, we might be selecting for the instances that are most straightforward to predict.
For XLM-R, the number of instances in this last setting is also low to very low, so those particular F1 scores may be unreliable.

The results show that XLM-RoBERTa keeps up in performance when it comes to characterizing implications. 
However, when it comes to identifying whether a comment may contain toxicity in the first place, or identifying spans of tokens corresponding to elements of the toxicity, GPT4o clearly outperforms XLM-RoBERTa.

\subsubsection{Implication-level Performance}
Here we investigate the performance of the models on the level of implications. 
We do this to get a better sense of how well the models perform at the full task, rather than just the individual variables.

Included in the implication-level performance is are also the variables that are annotated w.r.t. the earlier variables which characterize component parts of the implication.
This includes the stakeholder attitudes, as well as the \texttt{implSarcasm} and \texttt{implStereotype} variables.

For \texttt{implSarcasm} and \texttt{implStereotype}, we give F1 scores in \autoref{tab:implication_level_f1}.
In the cond-impl setting, GPT4o outperforms the Prior baseline on both of these variables.
XLM-R does as well, although with a slim margin.
For these variables, we do not include a `cond-tokens' evaluation setting, because there is no single \_Tokens variable that would be suitable.

\begin{table}
\small
\let\mc\multicolumn
\centering
\setlength{\tabcolsep}{3pt}
\caption{F1 scores for variables that describe the implication.}
\label{tab:implication_level_f1}
\begin{tabular}{@{}llrrrrrr@{}}
\toprule
& & \mc{3}{c}{{GPT4o}} & \mc{3}{c}{{RoBERTa}}  \\
\cmidrule{3-5} \cmidrule{6-8}
& & \mc{1}{c}{\rule{4pt}{0pt}F1\rule{4pt}{0pt}}  & \mc{1}{c}{Prior} & \mc{1}{c}{Supp.} 
  & \mc{1}{c}{\rule{4pt}{0pt}F1\rule{4pt}{0pt}}  & \mc{1}{c}{Prior} & \mc{1}{c}{Supp.} \\
\midrule
\multirow{2}{*}{prop.}
& implSarcasm    & \cc{CDDEF9}.55   & \cc{C7DAF9}.62       & 1169       & \cc{DDE8FB}.38   & \cc{C7DAF9}.62       & 1169  \\
& implStereotype & \cc{D7E4FB}.45   & \cc{CCDDF9}.57       & 1169       & \cc{E3EDFC}.31   & \cc{CCDDF9}.57       & 1169  \\
\midrule
\multirow{2}{*}{cond-impl} 
& implSarcasm    & \cc{B8CFF7}{.79} & \cc{C4D8F8}.65       & 628        & \cc{BBD2F7}{.75} & \cc{C3D7F8}.67       & 399   \\
& implStereotype & \cc{C5D8F8}{.64} & \cc{CDDEF9}.55       & 628        & \cc{C8DAF9}{.61} & \cc{CBDCF9}.58       & 399   \\
\bottomrule
\end{tabular}
\end{table}

\paragraph{Summary Sentence Accuracy}
We perform another evaluation for GPT4o where we count the number of samples for which it produces the same annotation as at least one of the annotators.
This evaluation is performed on the same subset of English sentences used for the disagreement analysis.

GPT4o's predictions are \textit{identical} to one of the annotators for 10 out of 47 samples. 
For an additional 13 samples, we see a \textit{match}, but with a small difference.
Differences are: 
GPT4o predicts only a subset of the temporalities that the annotator included (3 samples); 
GPT4o is imprecise in its prediction for \texttt{subjectGroup} (5 samples), for example predicting ``The group mentioned'' or ``Certain individuals''; 
GPT4o predicted \texttt{hasOther}, when annotators did not (2 samples);
or,
GPT4o predicted nature instead of behavior or vice versa (3 samples).
For 14 samples, GPT4o predicted \textit{no implication}.
Finally, for 10 samples, there was \textit{no match} with any annotator's implication.

\begin{table}
\small
\centering
\newcommand{\mycol}[1]{\multicolumn{1}{r}{#1}}
\caption{GPT4o stakeholder attitude evaluation. Rows corresponds to differences in ordinal values assigned by GPT4o vs. Human annotators. Row marked `\#' shows number of samples where that attitude was annotated.}
\label{tab:attitudes}
\begin{tabular}{@{}lrrrrrr@{}}
\toprule
  & \multicolumn{3}{c}{Author} &  \multicolumn{2}{c}{Typical} & Expert \\
\cmidrule(lr){2-4} \cmidrule(lr){5-6} \cmidrule(l){7-7}
  & \mycol{Belief} & \mycol{Prefer} & \mycol{Account} & \mycol{Belief} & \mycol{Prefer} & Belief \\
\#    & 20     &  9     &   1    & 21    &  8     & 17     \\
\midrule
0     & 45\%   &  0\%   &   0\%  & 19\%  & 25\%   & 12\%   \\
1     & 45\%   & 56\%   & 100\%  & 62\%  & 75\%   & 47\%   \\
2     &  5\%   & 33\%   &   0\%  & 19\%  &  0\%   & 41\%   \\
3     &  5\%   & 11\%   &   0\%  &  0\%  &  0\%   & 0\%    \\
\bottomrule
\end{tabular}
\end{table}

\paragraph{Attitudes}
To evaluate the stakeholder attitudes, we look at samples for which the summary sentences were either a \textit{match} or \textit{identical} to one of the annotators.
For each attitude, the annotators chose between the ordinal values ``Very low'', ``Low'', ``Medium'', ``High'', ``Very high'', or ``Not applicable''.
The last option is there to indicate when there is too little information in a comment and its context to tell the author's attitude.

In \autoref{tab:attitudes}, we report the number of GPT4o's predictions that were off by 0, 1, 2, or 3 steps in the ordinal scale.
What we can see is that \texttt{authorPrefer} is the attitude with which GPT4o struggles most.
For all attitudes, over half of the predictions are 0 or 1 steps away.

\subsection{Discussion}
%
The results show that toxic reasoning as a task, and the ToxiREX test set in particular, present a challenging problem.
Neither GPT4o nor the fine-tuned XLM-RoBERTa are full solutions to this problem.

XLM-RoBERTa generally underperforms GPT4o. 
This is expected as it was fine-tuned on labels produced by GPT4o.

For the part of the data we analyzed, GPT4o gives characterizations of toxic implications that match or are identical to one of the human annotations in 23 out of 47 cases.
In many of the remaining cases, GPT4o does not match a human's annotation.
However, looking at GPT4o's free-text description of the implication it (presumably) attempted to characterize, we can see that often those implications do still match (one of) the human annotations. 
This suggests that the model struggles with correctly using the toxic reasoning schema as presented in the Pydantic model.
However, for 14 out of 47 instances, GPT4o did not predict any implication.
Thus, the model still misses out on many of the toxic implications spotted by human annotators.
And, more deficiencies can be seen in the results on the attitudes and the span-level predictions. 

From the results it is clear that there is still much to improve with regards to models' ability to exhaustively list and characterize the toxic implications of social media comments.

\section{Conclusion}
We have released ToxiREX, a contextual, multilingual dataset annotated for toxic reasoning on Reddit comments, where annotations provide structured characterizations of what a comment implies. 
The annotations capture and explain implicit and context-dependent toxicity, and can be mapped to existing toxicity taxonomies.
It includes a test set of almost 3,000 comments, split between six languages, and annotated by native speakers.
The ToxiREX training set consists of roughly 125,000 comments with silver-standard labels generated by GPT4o. 
The dataset's conversational threads, often about controversial topics, include toxic implications of various kinds.
In the process of annotating the test set, we have shown how the toxic reasoning schema has, with few exceptions, proven expressive enough for the human annotators. 

To produce baseline results and to show how to evaluate models on the test set, we evaluate GPT4o and a fine-tuned XLM-RoBERTa.
We have shown how GPT4o-based annotation (and prediction) is made possible by using structured outputs.
When GPT4o is evaluated, its performance is strong for a zero-shot setup, but also shows that automatically detecting and characterizing toxic implications is still an open challenge. 

Both the toxic reasoning approach and the ToxiREX dataset present various opportunities for future work.
For example, the multilinguality of our dataset presents an opportunity to better understand toxic language across cultures.
Characterizing toxicity in different languages and different cultures using the same schema allows for direct comparisons.
Do different cultures appeal to different kinds of negative situations, behaviors, and qualties? 
Another direction is to pursue strategies that improve performance on the toxic reasoning tasks, i.e. detecting and characterizing toxic implications. 

For ToxiREX we used a `bottom-up' approach to construct our dataset.
We asked annotators to become familiar with the full toxic reasoning schema, and to annotate according to it.
Afterwards, we can map those annotations to existing taxonomies of toxic language.
In future work, we can also identify any instances that do not cleanly map onto existing categories, or identify if instances mapped to the same existing category still differ along some other axis, potentially giving cause to recognize new subcategories.

Further opportunities lie in `top-down' data annotations, where annotators select familiar categories of toxic language, and then verify that their understanding comports with how it maps onto the toxic reasoning schema.
For example, when an annotation of `Intimidation' is chosen, we ask if the author indeed desires someone be subjected to a negative situation, and if so, what spans correspond to the subject and the negative situation.
This would be substantially less labor-intensive than the bottom-up approach, which requires a full characterization of all aspects, rather than just those already indicated as relevant by the mapping. 
Such a top-down annotation campaign will be complementary by allowing a larger scale, in exchange for trading in some expressivity, as the categories will be fixed beforehand.


    \paragraph{Funding Statement} This research was supported by Huawei Finland through the DreamsLab project. All content represented the opinions of the authors, which were not necessarily shared or endorsed by their respective employers and/or sponsors.

    \paragraph{Competing Interests} 
    The authors declare none.

    \paragraph{Data Availability} 
    The ToxiREX dataset can be found on Github at \url{https://github.com/cltl/toxirex/}. 
    

    \paragraph{Ethics} 
    Ethical approval was not required.



\bibliographystyle{nlplike}
\bibliography{references}

\appendix
\section{Annotation Guidelines \& Questions}
\label{sec:annotation_questions}

\begin{lstlisting}[style=llmprompt]
Your task is to review a selection of Reddit comments. You will be asked a number of questions about each comment, focusing on that comment's potential toxicity.

We ask you to first read the context (see 'Reddit Thread').  The context consists of the following parts: 
 - a post title (shown at the top in a large font);
 - optionally, the text of the post itself; and finally, 
 - the comments section, where people are talking about the post.

After reading the context, please proceed to answer the questions about the current (highlighted in yellow) comment.  The current comment will be repeated as well, and will remain at the top of the page after you scroll past it. NOTE: The box that shows the current comment at the top of the page can be resized by dragging the bottom right corner.

The first few questions will determine if this comment is relevant (toxic in the right way). If it is not relevant you will be asked to continue to the next comment. If it is relevant, you will be asked what you think the main toxic implication is.

The rest of the questions should be answered with respect to that main implication. The main toxic implication you write down will also remain at the top of the page, underneath the current comment, in case you need to refer back to it.
\end{lstlisting}%

\begin{table}[h]
\setlength{\tabcolsep}{3pt}
\small
\caption{Annotation Questions}
\label{tab:annotation_questions}
\begin{subtable}{\textwidth}
\subcaption{Questions 1--7}
\begin{tabular}{@{}lp{7.5cm}l@{}}
\toprule
\# & Question \& Answer options & Variable
\\ \midrule
1 &
  \textbf{In your view, do think this comment might be (perceived as) potentially toxic, polarizing, or otherwise negative towards a group or individual?}
  --- 
  "Yes/Maybe", "No" &
  \multirow{9}{*}{toxicity} \\
2 &
  \textbf{Is it counter-speech? (An argument-based reply aimed at the previous comment, providing an alternative perspective.)}
  --- 
  \textsc{checkbox} &
   \\
- &
  \textit{If 'No' or 'No'+'Counter-speech', skip the following questions and go to the next comment.} 
   &
   \\
\midrule
3 &
  \textbf{Is the comment only toxic because it uses inappropriate words? e.g. use of slurs, name-calling, etc.}
  ---
  "Yes", "No" &
  justInappropriate \\
- &
  \textit{If 'Yes', skip the following questions and go to the next comment.} 
   &
   \\
\midrule
4\&5 &
  \textbf{If the toxicity stems from what the text is either explicitly or implicitly communicating, please write one sentence that conveys the main toxic implication.}
  ---
  \textsc{textbox} &
  implication \\
 &
  \textit{If not (i.e. the toxicity cannot be captured in an implication), choose 'Different kind of toxicity', and continue to the next comment.} 
  ---
  \textsc{checkbox} &
  hasImplication \\
\midrule
6 &
  \textbf{Who is the subject of the implication?}
  ---
    "the author themselves and/or their ingroup",
    "another participant in the conversation and/or the group they belong to",
    "an individual outside of the conversation",
    "another group",
    "none of the above" &
  subject \\
- &
  \textit{If 'none of the above', continue to the next comment.} 
   &
   \\
\midrule
7 &
  \textbf{If the subject is a group, or if the subject is a member of a group, which group?}
  ---
  \textsc{textbox} &
  subjectGroup \\
\bottomrule
\end{tabular}
\end{subtable}
\end{table}

\begin{table}[h]
\ContinuedFloat
\small
\setlength{\tabcolsep}{3pt}
\begin{subtable}{\textwidth}
\subcaption{Questions 8--15}
\begin{tabular}{@{}lp{7cm}l@{}}
\toprule
\# & Question \& Answer options & Variable \\
\midrule
8 &
  \textbf{What kind of group is it, i.e. what characteristic(s) define the group?} ---
    "Sexual orientation", "Gender", "Disability", "Race/Ethnicity", "Age", "Religion", 
    "Famous individual", "Political affiliation", "Social belief", "Body image", 
    "Addiction", "Socioeconomic status", "Profession", "Nationality", "other", 
    "Not applicable" &
  subjectGroupType \\
\midrule
9 &
  \textbf{Please mark tokens that are part of references to the subject, including references by name, by pronoun, and by description.}
  ---
  "None"  /  \textsc{tokens in comment} &
  subjectTokens \\
\midrule
10\&11 &
  \textbf{Is the subject implicitly or explicitly compared or contrasted to another person or group? If so, who?}
  ---
  "No other",
  "the author themselves and/or their ingroup",
  "another participant in the conversation and/or the group they belong to",
  "an individual outside of the conversation",
  "another group",
  "none of the above"
   &
  hasOther \& other \\
\midrule
12 &
  \textbf{If the other is a group, or if the other is a member of a group, which group?}
  ---
  \textsc{textbox}
   &
  otherGroup \\
\midrule
13 &
  \textbf{Please mark tokens that are part of references to the other, including references by name, by pronoun, and by description.}
  ---
  "None"  /  \textsc{tokens in comment} 
  &
  otherTokens \\
\midrule
14 &
  \textbf{What is the main implication about?}
  ---
    "(a) the subject's circumstances, living conditions, physical condition or health, general wellbeing, access to resources, etc.", 
    "... (a.1) some kind of harm coming to the subject", 
    "(b) the subject's nature, inherent qualities or abilities, etc.", 
    "... (b.1) dehumanisation of the subject", 
    "(c) the subject's choices/decisions, lifestyle, beliefs, etc.", 
    "(d) a non-specific comparison (does not fall under (a), (b), or (c)) between the subject and the other", 
    "(e) unclear or none of the above"  &
  implCategory \\
\midrule
15 &
  \textbf{Please mark tokens that were relevant to determining your answer for the previous question.}
  ---
  "None"  /  \textsc{tokens in comment} &
  implTokens \\
\bottomrule
\end{tabular}
\end{subtable}
\end{table}

\begin{table}[h]
\ContinuedFloat
\small
\setlength{\tabcolsep}{3pt}
\begin{subtable}{\textwidth}
\subcaption{Questions 16--25}
\begin{tabular}{@{}lp{8cm}l@{}}
\toprule 
\# & Question \& Answer options & Variable \\
\midrule
16 &
  \textbf{Does the implication say something positive, negative, or neither (neutral)?}
  ---
  "Positive", "Neutral", "Negative" &
  implPolarity \\
\midrule
17 &
  \textbf{In your view, does the implication play into a known stereotype?}
  ---
  "Yes", "No" &
  implStereotype \\
\midrule
18 &
  \textbf{Is the implication conveyed through sarcasm?}
  ---
  "Yes", "No" &
  implSarcasm \\
\\
\midrule
19 &
  \textbf{At what point in time is the implication meant to hold?}
  ---
  "Past", "Present", "Future" &
  implTemporality \\
\midrule
20 &
  \textbf{Based on the comment, to what degree does the author seem to believe that: SUMMARY\_SENTENCE?}
  ---
  "Not applicable", "Very low", "Low", "Medium", "High", "Very high" &
  authorBelief \\
\midrule
21 &
  \textbf{Based on the comment, to what degree does the author seem to wish (or prefer) that: SUMMARY\_SENTENCE?}
  ---
  "Not applicable", "Very low", "Low", "Medium", "High", "Very high" &
  authorPrefer \\
\midrule
22 &
  \textbf{Based on the comment, to what degree does the comment author seem personally committed to, or accountable/responsible for, the truth of the following statement: SUMMARY\_SENTENCE?}
  ---
  "Not applicable", "Very low", "Low", "Medium", "High", "Very high" &
  authorAccount \\
\midrule
23 &
  \textbf{In your view, to what degree would the general public believe that: SUMMARY\_SENTENCE?}
  ---
  "Not applicable", "Very low", "Low", "Medium", "High", "Very high" &
  typicalBelief \\
\midrule
24 &
  \textbf{In your view, to what degree would the general public wish (or prefer) that: SUMMARY\_SENTENCE?}
  ---
  "Not applicable", "Very low", "Low", "Medium", "High", "Very high" &
  typicalPrefer \\
\midrule
25 &
  \textbf{In your view, to what degree would an expert believe that: SUMMARY\_SENTENCE?}
  ---
  "Not applicable", "Very low", "Low", "Medium", "High", "Very high" &
  expertBelief \\
\bottomrule
\end{tabular}
\end{subtable}
\end{table}
\clearpage

\section{Summary sentence generation pseudocode} \label{apx:summary_sentence_pseudocode}

\begin{pseudocode*}[h]
\small
\lstset{basicstyle=\ttfamily\small, breaklines=true, frame=single}
\begin{lstlisting}[language=Python]
def summary_sentence(subject, subjectGroup, subjectTokens, hasOther, other, otherTokens, implTemporality, implPolarity, implTopic): 

  match subject:
    case Personal:    subjectStr = "An Individual"
    case GroupMember: subjectStr = subjectGroup + " and a related individual"
    case GroupWhole:  subjectStr = subjectGroup
  subjectStr += f" ({','.join(subjectTokens)}) "

  match other:
    case Personal:    otherStr = "An Individual"
    case GroupMember: otherStr = otherGroup + " and a related individual"
    case GroupWhole:  otherStr = otherGroup
  otherStr += f" ({','.join(otherTokens)})"

  verbStr = "had/" if Past in implTemporality else "./"
  isPresent = Present in implTemporality
  match (isPresent, other):
    case (true, Personal):  verbStr += "has/"
    case (true, _):         verbStr += "have/"
    case (false, _):        verbStr += "./" 
  verbStr += "will have" if Future in implTemporality else "."

  polarityStr = implPolarity.toString()

  match implTopic:
    case Situational: topicStr = "circumstances"
    case Harm:        topicStr = "harm come to them"
    case Qualitative: topicStr = "nature"
    case SubHuman:    topicStr = "lack of humanity"
    case Behavioral:  topicStr = "behavior"
    default:          topicStr = "property"

  comparisonStr = "compared to " + otherStr if hasOther else ""

  return f"{subjectStr} {verbStr} {polarityStr} {topicStr} {comparisonStr}"
  \end{lstlisting}
      \caption{Generation of summary sentence.}
      \label{pc:summary_sentence}
\end{pseudocode*}

\clearpage

\section{Annotator Information}
\label{apx:annotator_information}

\begin{table}[h]
\small
\centering
\renewcommand{\arraystretch}{1.2}%
\caption{Experience of annotators with toxic language. The scale for experience with toxicity is `None', `Little', `Average', `Extensive', or `Unsure'.}
\label{tab:annotator_experience}
\begin{tabular}{@{}l@{\hspace{1mm}}ll@{\hspace{10mm}}l@{\hspace{5mm}}l@{}}
\toprule
&       & \multicolumn{2}{l}{\textbf{Experience with Toxicity}} & \\
\cmidrule(r){3-4}
& \textbf{ID} & \textbf{Targeted By}& \textbf{Viewed}  & \textbf{Position / Program}    \\
\midrule
\multirow{3.55}{*}{\rotatebox{90}{|~~Dutch}}   
& D1    & None                      & Yes              & Staff         \\
& D2    & None                      & Yes              & Staff         \\
\cmidrule{2-5}
\multirow{2.8}{*}{\rotatebox{90}{English\hspace{3.8mm}}} 
& D3/E1 & None                      & Yes              & PhD Candidate \\
\cmidrule{2-5}
& E2    & -                         & -                & PhD Candidate \\
& E3    & None                      & Extensive        & Staff         \\
\midrule
\multirow{3.2}{*}{\rotatebox{90}{Spanish}}   
& S1    & None                      & Little           & BSc Artificial Intelligence    \\
& S2    & Extensive                 & Unsure           & BSc Artificial Intelligence    \\
& S3    & None                      & Average          & MSc Linguistics                \\
\midrule
\multirow{3.2}{*}{\rotatebox{90}{German}}    
& G1    & None                      & Little           & MA Humanities Research         \\
& G2    & Little                    & Little           & BSc Articifial Intelligence    \\
& G3    & Average                   & None             & BSc Articifial Intelligence    \\
\midrule
\multirow{3.2}{*}{\rotatebox{90}{Turkish}}   
& T1    & None                      & Little           & BSc Articifial Intelligence    \\
& T2    & Extensive                 & Extensive        & BSc Articifial Intelligence    \\
& T3    & None                      & None             & BSc Articifial Intelligence    \\
\midrule
\multirow{3.2}{*}{\rotatebox{90}{Arabic}}    
& A1    & None                      & Unsure           & BSc Articifial Intelligence    \\
& A2    & None                      & Average          & BSc Computer Science           \\
& A3    & Little                    & Extensive        & BSc Computer Science           \\
\bottomrule
\end{tabular}
\end{table}

\begin{table}
\small
\centering
\setlength{\tabcolsep}{1.8pt}
\caption{Demographic information of annotators. One person annotated both Dutch and English (D3/E1).}
\label{tab:annotator_info_A}
\begin{tabular}{@{}lllllllll@{}}
\toprule
 & \textbf{ID} & \textbf{Age} & \textbf{Gender} & \textbf{Country} & \textbf{Native Language} & \textbf{Political Leaning} & \textbf{Religion} & \textbf{Sexuality} \\
\midrule
\multirow{3.2}{3mm}{\rotatebox{90}{|~~Dutch~~}}
 & D1    & 60-66 & Female & Netherlands & Dutch             & Left                     & Agnostic         & Heterosexual \\
 & D2    & 60-66 & Male   & Netherlands & Dutch             & Leftwing Liberal         & Atheist          & Heterosexual \\
\cmidrule{2-9}
\multirow{3.2}{3mm}{\rotatebox{90}{English\hspace{3.8mm}}}
 & D3/E1 & 25-31 & Male   & Netherlands & Dutch             & Social Liberal           & Atheist          & Heterosexual \\
\cmidrule{2-9}
 & E2    & 25-31 & Female & Iran        & Persian           & -                        & -                & -            \\
 & E3    & 39-45 & Male   & Russia      & Russian           & Liberal                  & Atheist          & Heterosexual \\
\midrule
\multirow{3}{3mm}{\rotatebox{90}{Spanish}}
 & S1    & 25-31 & Male   & Mexico      & English, Spanish  & Conservative leaning     & Agnostic         & Heterosexual \\
 & S2    & 18-24 & Female & Venezuela   & Spanish           & Libertarian              & Atheist          & Bisexual     \\
 & S3    & 25-31 & Female & Greece      & Greek, Spanish    & Socialist                & Agnostic Atheist & Heterosexual \\
\midrule
\multirow{3}{3mm}{\rotatebox{90}{German}}
 & G1    & 25-31 & Female & Germany     & German            & Left                     & Agnostic         & Queer        \\
 & G2    & 25-31 & Female & Germany     & Russian, German   & Left                     & None             & Heterosexual \\
 & G3    & 18-24 & Male   & Germany     & German            & Progressive conservatism & Atheist          & Heterosexual \\
\midrule
\multirow{3}{3mm}{\rotatebox{90}{Turkish}}
 & T1    & 18-24 & Male   & Turkey      & Turkish           & Liberal                  & Islam            & Heterosexual \\
 & T2    & 18-24 & Male   & Turkey      & English, Turkish  & Non-political            & Islam            & Heterosexual \\
 & T3    & 18-24 & Male   & Turkey      & Turkish           & Left-wing \& libertarian & Islam            & Heterosexual \\
\midrule
\multirow{3}{3mm}{\rotatebox{90}{Arabic}}
 & A1    & 18-24 & Female & Morocco     & Arabic            & None                     & Islam            & Heterosexual \\
 & A2    & 18-24 & Female & Egypt       & Arabic            & Moderate                 & Islam            & Heterosexual \\
 & A3    & 18-24 & Male   & Jordan      & Arabic            & Moderate / Centrist      & Islam            & Heterosexual \\
\bottomrule
\end{tabular}
\end{table}

\begin{table}[h]
\small
\centering%
\let\mc\multicolumn%
\caption{The nr. of comments for which $N \in \{0,1,2,3\}$ annotators chose the option which continues the annotation, by language.}
\label{tab:nr_of_continuing_annotators}
\begin{tabular}{@{}lrrrrrrr@{}}
\toprule
 & \mc{1}{l}{$N$} & \mc{1}{l}{\textbf{AR}} & \mc{1}{l}{\textbf{DE}} & \mc{1}{l}{\textbf{EN}} & \mc{1}{l}{\textbf{ES}} & \mc{1}{l}{\textbf{NL}} & \mc{1}{l}{\textbf{TR}} \\ \midrule
toxicity            & 0 & 192 & 127 & 189 & 147 & 160 & 164 \\
                    & 1 & 101 & 119 & 116 &  85 & 123 &  69 \\
                    & 2 & 114 & 140 & 123 &  89 &  84 &  64 \\
                    & 3 &  99 & 120 &  92 & 174 &  52 & 153 \\
justInappropriate   & 0 & 216 & 151 & 188 & 172 & 194 & 282 \\
                    & 1 & 135 & 153 & 148 & 109 & 130 & 104 \\
                    & 2 & 109 & 129 & 115 & 106 &  70 &  49 \\
                    & 3 &  46 &  73 &  69 & 108 &  25 &  15 \\
hasImplication      & 0 & 337 & 340 & 254 & 207 & 239 & 331 \\
                    & 1 &  84 & 100 & 132 & 146 &  90 &  61 \\
                    & 2 &  57 &  45 &  87 &  80 &  61 &  39 \\
                    & 3 &  28 &  21 &  47 &  62 &  29 &  19 \\ 
\bottomrule
\end{tabular}
\end{table}

\clearpage

\section{Early version of Toxic Reasoning Schema (TRS0)} \label{apx:trs0}
\normalsize
Here we describe the traits that make up an older version (TRS0) of the toxic reasoning schema. 
It is very similar to the latest version, but less complete, and had a less general way of categorizing the implications. 
See \autoref{tab:trs0_mapping} for how we envisioned mapping this version of the schema to the IHC \citep{elsherief_latent_2021}.

The schema's first trait captures the \textbf{implication}(s) of a text. 
It characterizes the proposition that is central to the text's implication. 
We identify three high-level categories of {implications}:\footnote{By implication we mean the property (of a single group) or relation (between groups) that were either implicitly or explicitly applied to a group or groups.}%
\begin{itemize}%
    \item{}[Negative/Positive]Condition(\textit{group}): a condition (e.g. environment, circumstance, etc.) has, does, or will apply to \textit{group}.  
    \item{}[Negative/Positive]Quality(\textit{group}): a \textit{group} or a \textit{group}-member possesses a particular quality;
    \item Inferior(\textit{group1}, \textit{group2}) /\\
    Superior(\textit{group2}, \textit{group1}): \textit{group1} is inferior in some way to \textit{group2}.
\end{itemize}%
The first category contains statements about a group being affected by their environment. 
For example, \textit{being evicted} would be a negative condition.
The second category is about the group's inherent qualities, or nature. 
This could include statements describing groups of people as `sub-human' or `vermin'.
Finally, the third category involves how groups relate to each other, e.g. suggesting that people of one ethnicity are more intelligent than another.



\begin{table}
    \centering
    \small
    \let\mr\multirow     %
    \let\mc\multicolumn  
    \newcommand{\ri}[2]{\multicolumn{#1}{c|}{#2}}
    \newcommand{\dt}{\color{gray}$\cdot$}
    \newcommand{\cml}{\dt}
    \newcommand{\gr}{\color{gray}}
    \newcommand{\ra}{\quad}
    \newcommand{\str}{\hphantom{aaa}$\ast$\hphantom{aaa}}
    \let\cm\checkmark
    \let\q\quad
    \footnotesize
    \setlength\tabcolsep{2pt}
    \begin{tabular}{llccccccccc}
\toprule
         &
         & \ri{1}{\mr{2}{*}{\shortstack{Threat\, /\\Intimidation}}}
         & \ri{1}{\mr{2}{*}{Incitement}}
         & \ri{1}{\mr{2}{*}{Grievance}}
         & \ri{2}{Inferiority}
         & \mc{2}{c}{Misinformation}
\\ 
         &&\ri{1}{}&\ri{1}{\;}&\ri{1}{\;}&\ri{1}{\str}&\ri{1}{Dehuman.}&\ri{1}{\str}&Stereotype
\\ \midrule
    \mr{12}{*}{\rotatebox{90}{Implications}}
         & Something(\textit{group})                     & \dt & \dt & \dt & \dt & \dt & \cm & \dt \\
         & \q SomethingNegative(\textit{group})          & \dt & \dt & \dt & \dt & \dt & \cml& \dt \\
         & \q \q NegativeCircumstance(\textit{group})       & \dt & \dt & \dt & \dt & \dt & \cml& \dt \\
         & \q \q \ra Harmed(\textit{target})              & \cm & \cm & \dt & \dt & \dt & \cml& \dt \\
         & \q \q \ra Harmed(\textit{in\_group})           & \dt & \dt & \cm & \dt & \dt & \cml& \dt \\
         & \q \q NegativeNature(\textit{group})         & \dt & \dt & \dt & \dt & \dt & \cml& \dt \\
         & \q \q \ra NotHuman(\textit{target})            & \dt & \dt & \dt & \cml& \cm & \cml& \dt \\
         & \q \ra Inferior(\textit{target}, \textit{in\_group})  & \dt & \dt & \dt & \cm & \dt & \cml& \dt \\
         & \q SomethingPositive(\textit{group})          & \dt & \dt & \dt & \dt & \dt & \cml& \dt \\
         & \q \ra Superior(\textit{in\_group}, \textit{target})  & \dt & \dt & \dt & \cm & \dt & \cml& \dt \\
         & \ra KnownStereotype(\textit{target})           & \dt & \dt & \dt & \dt & \dt & \cml& \cm \\
\midrule
         & temporality                            & future 
                                                        & future
                                                              & \dt & \dt & \dt & \dt & \dt \\
\midrule
    \mr{6}{*}{\rotatebox{90}{Attitudes}}
         & \textsl{author\_belief}                  & +   & \dt & +   & +   & +   & +   & \dt \\ 
         & \textsl{author\_preference}              & \dt & +   & $-$ & \dt & \dt & \dt & \dt \\ 
         & \textsl{author\_accountability}          & +   & $-$ & $-$ & \dt & \dt & \dt & \dt \\ 
         & \textsl{typical\_belief}                 & \dt & \dt & $-$ & \dt & \dt & \dt & \dt \\ 
         & \textsl{typical\_preference}             & \dt & \dt & $-$ & \dt & \dt & \dt & \dt \\ 
         & \textsl{expert\_belief}                  & \dt & \dt & $-$ & \dt & \dt & $-$ & $-$ \\ 
\bottomrule
    \end{tabular}
    \caption{Mapping between TRS0 and the IHC \citep{} (sub-)classes. Each column represents an IHC (sub-)class, it contains: (1) check marks for the implications that could be involved in a sample of that class; (2) the values for the temporality, specificity, and origin that are required by that class (or `$\cdot$' for no requirements); and (3) constraints on the possible values for the attitudinal attributes for that class (with `+' indicating a minimum constraint: `$> 0.5$', `$-$' indicating a maximum constraint: `$< 0.5$', and `$\cdot$' indicating no constraints). }
    \label{tab:trs0_mapping}
\end{table}

\clearpage

\section{ToxiREX Events}\label{apx:toxirex_events}

\begin{table}[h]%
\small%
\caption{Events to which sub-threads in ToxiREX are linked. The training data includes sub-threads for each event. Whether an event is part of the test data is indicated in the `Test?' column.}%
\begin{subtable}{0.46\textwidth}%
\begin{tabular}{@{}lll@{}}%
\toprule%
\textbf{Event Title}                                                     & \textbf{Test?} \\ \midrule
2022\_European\_heatwaves                                                & Yes               \\
2022\_FIFA\_World\_Cup                                                   & Yes               \\
2023\_Israel–Hamas\_war                                                  & Yes               \\
2023\_Turkey–Syria\_earthquakes                                          & Yes               \\
Baldur's\_Gate\_3                                                        & No                \\
Chained\_Echoes                                                          & No                \\
Chicory:\_A\_Colorful\_Tale                                              & No                \\
Cocoon                                                                   & No                \\
COVID-19\_pandemic                                                       & Yes               \\
Cuphead\_in\_the\_Delicious\_Last\_Course                                & No                \\
Cyberpunk\_2077:\_Phantom\_Liberty                                       & No                \\
Dave\_the\_Diver                                                         & No                \\
Death\_and\_state\_funeral\_of\_Elizabeth\_II                            & Yes               \\
Diablo\_IV                                                               & No                \\
Disco\_Elysium:\_The\_Final\_Cut                                         & Yes               \\
Dwarf\_Fortress                                                          & Yes               \\
Elden\_Ring                                                              & Yes               \\
Eurovision\_Song\_Contest\_2021                                          & No                \\
Eurovision\_Song\_Contest\_2022                                          & No                \\
Eurovision\_Song\_Contest\_2023                                          & No                \\
Fall\_of\_Kabul\_(2021)                                                  & Yes               \\
Final\_Fantasy\_XIV:\_Endwalker                                          & Yes               \\
Forza\_Horizon\_5                                                        & Yes               \\
God\_of\_War                                                             & Yes               \\
God\_of\_War\_Ragnarök                                                   & No                \\
Hades                                                                    & No                \\
Honda\_car\_recall\_2021                                                 & No                \\
Honda\_car\_recall\_2023                                                 & No                \\
Hyundai\_car\_recall\_2021                                               & Yes               \\
Hyundai\_car\_recall\_2023                                               & No                \\
\bottomrule
\end{tabular}%
\end{subtable}%
%
\begin{subtable}{0.54\textwidth}%
\begin{tabular}{@{}lll@{}}%
\toprule%
\textbf{Event Title}                                                     & \textbf{Test?} \\ \midrule
Jack\_Jeanne                                                             & No                \\
January\_6\_United\_States\_Capitol\_attack                              & Yes               \\
Jeep\_car\_recall\_2023                                                  & Yes               \\
Kia\_car\_recall\_2023                                                   & Yes               \\
Mahsa\_Amini\_protests                                                   & Yes               \\
Mass\_Effect\_Legendary\_Edition                                         & Yes               \\
Metroid\_Prime\_Remastered                                               & No                \\
Microsoft\_Flight\_Simulator                                             & Yes               \\
Neon\_White                                                              & No                \\
Persona\_5\_Royal                                                        & No                \\
Portal\_Companion\_Collection                                            & No                \\
Psychonauts\_2                                                           & Yes               \\
Quake\_II\_-\_Enhanced\_Edition                                          & No                \\
Resident\_Evil\_4                                                        & No                \\
Rogue\_Legacy\_2                                                         & Yes               \\
Russian\_invasion\_of\_Ukraine                                           & Yes               \\
Sea\_of\_Stars                                                           & Yes               \\
Spider-Man\_2                                                            & No                \\
Super\_Mario\_Bros.\_Wonder                                              & No                \\
Tesla\_car\_recall\_2023                                                 & No                \\
Tetris\_Effect:\_Connected                                               & Yes               \\
The\_House\_in\_Fata\_Morgana\_     \\
      \multicolumn{1}{c}{-\_Dreams\_of\_the\_Revenants\_Edition}         & Yes               \\
The\_Legend\_of\_Heroes:\_Trails\_from\_Zero                             & No                \\
The\_Legend\_of\_Zelda:\_Tears\_of\_the\_Kingdom                         & Yes               \\
The\_Stanley\_Parable:\_Ultra\_Deluxe                                    & Yes               \\
The\_Witcher\_3:\_Wild\_Hunt\_-\_Complete\_Edition                       & No                \\
Tony\_Hawk's\_Pro\_Skater\_1\_+\_2                                       & No                \\
Turbo\_Overkill                                                          & No                \\
Volkswagen\_car\_recall\_2021                                            & No                \\
Walkabout\_Mini\_Golf                                                    & No                \\
Xenoblade\_Chronicles\_3:  \\
\multicolumn{1}{r}{\_Expansion\_Pass\_Wave\_4\_-\_Future\_Redeemed}      & Yes               \\ 
\bottomrule
\end{tabular}
\end{subtable}
\end{table}

\clearpage

\section{Pydantic Model TRS0} \label{apx:pydantic_trs0}

\lstset{
  language=Python,
  basicstyle=\ttfamily\footnotesize,
  keywordstyle=\color{blue},
  commentstyle=\color{gray},
  stringstyle=\color{teal},
  breaklines=true,           
  breakatwhitespace=false,   
  columns=flexible,          
  keepspaces=true,           
  showstringspaces=false     
}

\begin{lstlisting}
class Property(str, Enum):
    harmed = "Harmed(GROUP_A)"
    not_human = "NotHuman(GROUP_A)"
    known_stereotype = "KnownStereotype(GROUP_A)"
    inferior = "Inferior(GROUP_A, GROUP_B)"
    superior = "Superior(GROUP_B, GROUP_A)"
    something_positive = "SomethingPositive(GROUP_A)"
    something_negative = "SomethingNegative(GROUP_A)"
    nothing_implied = "SomethingNeutral(GROUP_A)"

class Temporality(str, Enum):
    past = "past"
    present = "present"
    future = "future"

class ToxicReasoning(BaseModel):
    message_nr: int = Field(description="The number of the message in the thread to which the reasoning applies.")

    implication: str = Field(description="If the message seems to imply something about a person or a group, give a short description. If there is no clear implication do not generate a ToxicReasoning.")

    prop: Property = Field(title="Property", description="The property that the author is implying has, will, or does apply to a group. Choose the most specific property that clearly applies.")
    prop_span: str = Field(title="Property span", description="The span in the original text most indicative of the property being implied.")

    group_a_description: str = Field(title="GROUP_A", description="The group about whom the property is asserted, or otherwise the first of two.")
    group_b_description: str = Field(title="GROUP_B", description="An optional second group of which something is implied, possibly used as a point of comparison to GROUP_A.")
    group_a_span: str = Field(description="The span in the original text most indicative of GROUP_A.")
    group_b_span: str = Field(description="The span in the original text most indicative of GROUP_B.")

    when: Temporality = Field(description="At what point in time the property is implied to apply to (one of) the groups.")

    author_belief: float = Field(description="[0-1]: Probability that the author believes their implication.")
    author_preference: float = Field(description="[0-1]: Probability that the author would prefer for their implication to be true.")
    author_responsibility: float = Field(description="[0-1]: Probability that the author is or feels personally responsible for the (future) truth of their implication.")
    typical_belief: float = Field(description="[0-1]: Probability that ordinary people would believe the author's implication.")
    typical_preference: float = Field(description="[0-1]: Probability that ordinary people would prefer for the implication to be true.")
    expert_belief: float = Field(description="[0-1]: Probability that experts would believe the author's implication.")

    confidence: float = Field(description="[0-1]: Confidence that the above is correct.")


class ThreadReasonings(BaseModel):
    message_reasonings: list[ToxicReasoning] \
        = Field(description="A list of toxic reasonings describing something said or implied in one of the messages in the thread.")
\end{lstlisting}
\clearpage

\section{Pydantic Model TRS1} \label{apx:pydantic_trs1}
\begin{lstlisting}
class GroupRole(str, Enum):
    author = "the author themselves and/or their ingroup"
    participant = "another participant in the conversation and/or the group they belong to"
    outside = "an individual outside of the conversation"
    another = "another group"
    na = "none of the above"


class GroupCharacteristic(str, Enum):
    sexual = "Sexual orientation"
    gender = "Gender"
    disability = "Disability"
    race = "Race/Ethnicity"
    age = "Age"
    religion = "Religion"
    famous = "Famous individual"
    political = "Political affiliation"
    social = "Social belief"
    body_image = "Body image"
    addiction = "Addiction"
    socioeconomic = "Socioeconomic status"
    profession = "Profession"
    nationality = "Nationality"
    other = "Other"
    na = "Not applicable"


class ImplicationCategory(str, Enum):
    a = ("the subject's circumstances, living conditions, physical condition or health, general wellbeing, access to "
         "resources, etc.")
    a1 = "some kind of harm coming to the subject"
    b = "the subject's nature, inherent qualities or abilities, etc."
    b1 = "dehumanisation of the subject"
    c = "the subject's choices/decisions, lifestyle, beliefs, etc."
    d = "a non-specific comparison (does not fall under other categories) between the subject and the other"
    e = "unclear or none of the above"


class Polarity(str, Enum):
    positive = "Positive"
    neutral = "Neutral"
    negative = "Negative"


class Temporality(str, Enum):
    past = "Past"
    present = "Present"
    future = "Future"


class ToxicReasoning(BaseModel):
  # short description of implication
  implication: str = Field(description="If the message might imply something toxic about a person or a group, give the main implication as a single sentence. Do not describe the implication, but simply make explicit that which is implied; i.e. do not start with 'The comment/author implies ...', instead the sentence should generally start with the subject and continue with what is implied about the subject. If there is no clear implication do not generate a ToxicReasoning.")
  
  # SUBJECT
  subject_descr: str = Field(description="The subject (person or group) of the implication.")
  subject_role: GroupRole = Field(description="The role of the subject in the context of this thread.")
  subject_span: str = Field(description="The span in the original text most indicative of the subject.")
  subject_characteristic: GroupCharacteristic = Field(description="The characteristic that defines the subject group (if the subject is a group).")
  
  # OTHER
  has_other: bool = Field(description="If there is an 'other' to which the subject is (explicitly or implicitly) compared.")
  other_descr: str = Field(description="A description of the optional 'other' person or group.")
  other_role: GroupRole = Field(description="Who the other is in the context of this thread.")
  other_span: str = Field(description="The span in the original text most indicative of the other.")
  
  # IMPLICATION
  category: ImplicationCategory = Field(description="The category (topic) of the implication. Choose the most specific option that clearly applies.")
  impl_span: str = Field(description="The span in the original text most indicative of the Implication Category.")
  polarity: Polarity = Field(description="If positive, the main implication being true should be something for the subject to be happy about, or proud of. If negative, the main implication being true should be something for the subject to be sad about, or ashamed of.")
  stereotype: bool = Field(description="If the implication plays into a widely known stereotype.")
  sarcasm: bool = Field(description="If the implication is conveyed through sarcasm.")
  when: list[Temporality] = Field(description="At what point(s) in time the implication is meant to hold (at least 1).")
  
  # STAKEHOLDER ATTITUDES
  author_belief: float = Field(description="[0-1]: Probability that the author believes what their implication says was/is/will (or should [have] be[en]) the case.")
  author_preference: float = Field(description="[0-1]: Probability that the author prefers what their implication says was/is/will (or should [have] be[en]) the case.")
  author_responsibility: float = Field(description="[0-1]: Probability that the author is or feels personally responsible for the (past, present, or future) truth of what their implication says was/is/will (or should [have] be[en]) the case.")
  typical_belief: float = Field(description="[0-1]: Probability that ordinary people would believe what the implication says was/is/will (or should [have] be[en]) the case.")
  typical_preference: float = Field(description="[0-1]: Probability that ordinary people would prefer what the implication says was/is/will (or should [have] be[en]) the case.")
  expert_belief: float = Field(description="[0-1]: Probability that experts would believe what the implication says was/is/will (or should [have] be[en]) the case.")

  
class Toxicity(str, Enum):
    yes = "Yes/Maybe"
    no = "No"


class CommentAnnotation(BaseModel):
    message_nr: int = Field(description="The number of the message in the thread to which this annotation applies.")

    is_toxic: Toxicity = Field(description="If this comment might be perceived as toxic.")
    is_only_innapropriate: bool = Field(description="If the comment is only toxic because of an inappropriate or toxic word (rather than what is being said/implied).")
    is_counter_speech: bool = Field(description="If the comment is an argument-based counter of the previous comment, providing an alternative perspective.")
    toxic_reasoning: Optional[ToxicReasoning] = Field(
        description="If toxic and the toxicity stems from what the text is either explicitly or implicitly communicating this should contain a toxic reasoning. If toxic for a different reason, it should be null."
    )


class ThreadReasonings(BaseModel):
    comment_annotations: list[CommentAnnotation] = Field(
        description="A list of comment annotations specifying message toxicity."
    )
\end{lstlisting}

\clearpage

\section{Alternative Fine-tuning Strategy} \label{apx:decoder_strategy}
We explored an alternative fine-tuning strategy where a language model is expected to produce an instance of the pydantic model in the same way that GPT4o does.

We prepared supervised learning examples by adding sub-threads to a prompt and appending the silver-standard annotations of GPT4o.
The annotations were presented as instances of the Pydantic model converted to JSON format, with the JSON schema of the pydantic model being prepended to the prompt.
Thus, the model was expected to take its instructions from the JSON schema, and then produce a JSON output which adheres to it.
During inference, we used structured outputs \citep{willard_efficient_2023} to force the model to produce valid instances of the schema. 
We used Gemma-3-270m-it \citep{team_gemma_2025}, DeepSeek-Qwen-2.5-1.5b \citep{guo_deepseek-r1_2025}, and GPT-OSS-20b (in 4-bit precision) \citep{openai_gpt-oss-120b_2025}. 
These models were trained with Low-rank Adaption (LoRA) \citep{hu_lora_2022} parameter-efficient fine-tuning. 
We also tried Gemma-3-270m-it model with ordinary all-parameter fine-tuning.

Unfortunately, these preliminary experiments did not result in a model with better performance than the baseline presented in the main text.
We expect that a strategy like this one could still succeed. 
However, since our primary goal was to produce a simple baseline and our preliminary experiments did not succeed, we leave this for future work.

\label{lastpage}

\end{document}